\documentclass[lettersize,journal]{IEEEtran}

\usepackage[nocompress]{cite}

\usepackage{amsmath,amsfonts}
\usepackage{array}
\usepackage[caption=false,font=normalsize,labelfont=sf,textfont=sf]{subfig}
\usepackage{textcomp}
\usepackage{stfloats}
\usepackage{url}
\usepackage{verbatim}
\usepackage{graphicx}
\usepackage{cite}
\usepackage{ragged2e}
\usepackage[pagebackref=true,breaklinks=true,letterpaper=true,colorlinks,bookmarks=false]{hyperref}

\newcommand{\compat}{\textsc{3DCoMPaT\textsuperscript{++}}\xspace}
\newcommand{\papertitle}{\compat: An improved Large-scale 3D Vision Dataset for Compositional Recognition}

\newcommand{\nshapes}{10000\xspace}
\newcommand{\nshapesshort}{10K\xspace}
\newcommand{\nshapeclasses}{42\xspace}

\newcommand{\nmaterialclasses}{293\xspace}
\newcommand{\nmaterialclassescoarse}{13\xspace}

\newcommand{\npartsfine}{275\xspace}

\newcommand{\npartinstances}{9.86\xspace}

\newcommand{\npartscoarse}{43\xspace}

\newcommand{\avgncomposition}{1000\xspace}

\newcommand{\nstylespershape}{1000\xspace}
\newcommand{\nstylizedshapes}{10 million\xspace} 
\newcommand{\nstylizedshapesshort}{10M\xspace} 

\newcommand{\nviewpermodel}{8\xspace} 
\newcommand{\nviewpertype}{4\xspace} 
\newcommand{\nviews}{160 million\xspace} 
\newcommand{\nviewsshort}{160M\xspace}

\newcommand{\semipoint}[1]{\textbf{{(#1)}}}
\newcommand{\fpoint}[1]{\noindent \textbf{#1}}
\newcommand{\point}[1]{\vspace{-0.5em} \noindent \textbf{#1}}

\usepackage{url}
\usepackage{xspace}
\usepackage{subfig}
\usepackage{fontawesome5}

\usepackage{booktabs}
\usepackage{multirow}
\usepackage[table,xcdraw,svgnames]{xcolor}

\usepackage{algorithm}%
\usepackage{algorithmicx}%
\usepackage[noend]{algpseudocode}
\algrenewcommand\alglinenumber[1]{\tiny #1.}
\usepackage{adjustbox}

\usepackage{stmaryrd}
\usepackage{svg}
\usepackage{dsfont}

\definecolor{mygrey}{HTML}{dedede} 

\newcommand{\nodata}{\textcolor{mygrey}{\faIcon{times-circle}}\xspace}
\newcommand{\noanno}{\textcolor{LightSteelBlue}{\faIcon{question-circle}}\xspace}
\newcommand{\icoyes}{\textcolor{ForestGreen}{\faIcon{check-circle}}\xspace}
\newcommand{\icono}{\textcolor{IndianRed}{\faIcon{times-circle}}\xspace}

\begin{document}

\title{\papertitle}
\author{Habib~Slim,
        Xiang~Li,
        Yuchen~Li,
        Mahmoud~Ahmed,
        Mohamed~Ayman,
        Ujjwal~Upadhyay,\\
        Ahmed~Abdelreheem,
        Arpit~Prajapati,
        Suhail~Pothigara,
        Peter~Wonka,~\IEEEmembership{Senior Member,~IEEE,}
        and Mohamed~Elhoseiny,~\IEEEmembership{Senior Member,~IEEE}
\IEEEcompsocitemizethanks{
    \IEEEcompsocthanksitem Corresponding authors: H. Slim and M Elhoseiny with the Department of Computer Science, KAUST, Thuwal, Saudi Arabia. 
    \protect\\
    E-mail: habib.slim@kaust.edu.sa; mohamed.elhoseiny@kaust.edu.sa
    \IEEEcompsocthanksitem A. Prajapati, S. Pothigara are with Polynine, San Francisco, California. \protect
    \IEEEcompsocthanksitem X. Li, Y. Li, M. Ahmed, M. Ayman, U. Upadhyay, A. Abdelreheem, P. Wonka are with the Department of Computer Science, KAUST, Thuwal, Saudi Arabia. \protect\\}%
}

\IEEEtitleabstractindextext{%
\begin{abstract}
  \justifying
  In this work, we present \compat, a multimodal 2D/3D dataset with \nviews rendered views of more than \nstylizedshapes stylized 3D shapes carefully annotated at the part-instance level, alongside matching RGB point clouds, 3D textured meshes, depth maps, and segmentation masks. 
  \compat covers \nshapeclasses shape categories, \npartsfine fine-grained part categories, and \nmaterialclasses fine-grained material classes that can be compositionally applied to parts of 3D objects. %
  We render a subset of one million stylized shapes from four equally spaced views as well as four randomized views, leading to a total of \nviews renderings.
  Parts are segmented at the instance level, with coarse-grained and fine-grained semantic levels.
  We introduce a new task, called Grounded CoMPaT Recognition (GCR), to collectively recognize and ground compositions of materials on parts of 3D objects.
  Additionally, we report the outcomes of a data challenge organized at the CVPR conference, showcasing the winning method's utilization of a modified PointNet++ model trained on 6D inputs, and exploring alternative techniques for GCR enhancement.
  We hope our work will help ease future research on compositional 3D Vision. 
  The dataset and code have been made publicly available at \url{https://3dcompat-dataset.org/v2/}.
\end{abstract}

}

\maketitle 

\IEEEdisplaynontitleabstractindextext

\IEEEpeerreviewmaketitle
\vspace{3em}
\IEEEraisesectionheading{
  \section{Introduction}
  \label{sec:introduction}
}
  \begin{figure*}
    \centering
    \vspace{-2em}
    \includegraphics[width=\linewidth]{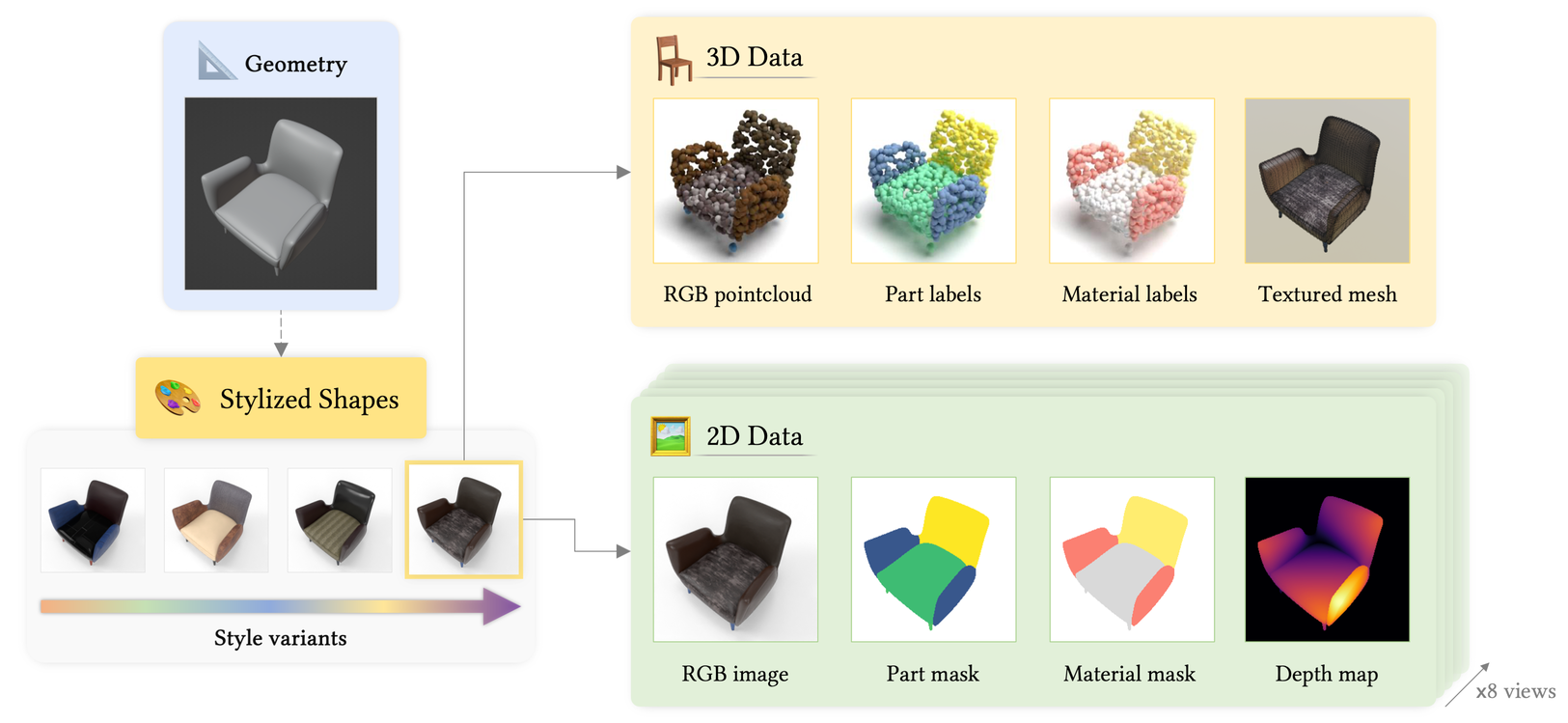}
    \vspace{-1em}
    \caption{\textbf{Data provided for each stylized shape.} For 3D: RGB pointclouds, textured shapes, and point-wise/triangle-wise part labels and material labels. For 2D: RGB images, depth maps, and corresponding part masks and material masks. Part and material annotations in 2D and 3D are provided in both \textit{coarse} and \textit{fine} semantic levels.
    In Figure~\ref*{fig:style_variants} of the appendix, we show additional style variants for various shapes.
    }
    \label{fig:all_modalities}
    \vspace{-0.1em}
  \end{figure*}

  \begin{figure}
    \centering
    \includegraphics[width=\linewidth]{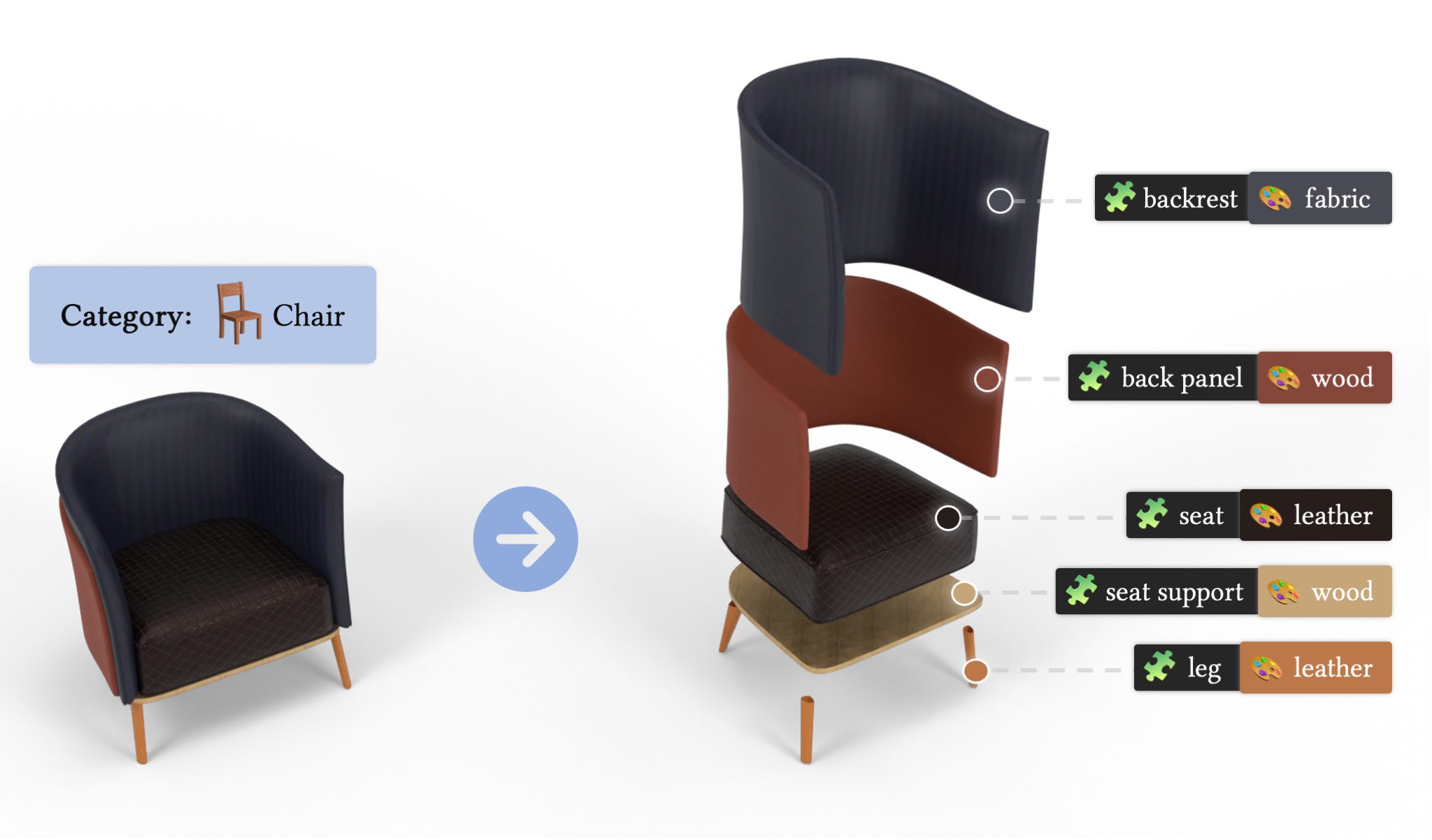}
    \caption{\textbf{Grounded CoMPaT Recognition (GCR).} Given an input shape, here: a chair, the task consists of \semipoint{a} recognizing the shape category and \semipoint{b} segmenting the part-material pairs composing it. }
    \label{fig:gcr_task}
    \vspace{-1em}
  \end{figure}

  \IEEEPARstart{M}{ultiple} datasets have been proposed to facilitate 3D visual understanding including ShapeNet~\cite{chang_shapenet_2015}, ModelNet~\cite{wu_modelnet_2015}, and PartNet~\cite{mo_partnet_2018}.
  High-quality datasets like OmniObject3D~\cite{wu_omniobject3d_2023} and ABO~\cite{collins_abo_2022} were introduced in an attempt to provide 3D assets with high-resolution, realistic textures.
  3D-Future~\cite{fu_3dfuture_2020} was also proposed and contains 10K industrial 3D CAD shapes of furniture with textures developed by professional designers.
  More recently, Objaverse~\cite{deitke_objaverse_2022} and its larger counterpart Objaverse-XL~\cite{deitke_objaverse-xl_2023} were introduced, which contain more than 10 million artist-designed 3D objects with high-quality textures.
  Despite these notable efforts to advance 3D understanding, recent object-centric 3D datasets (e.g.~\cite{chang_shapenet_2015, wu_modelnet_2015, deitke_objaverse_2022, deitke_objaverse-xl_2023}) and 3D scene datasets (e.g.~\cite{dai_scannet_2017, chang_matterport3d_2017}) lack part-level annotations.
  ShapeNet-Part~\cite{yi_shapenet_part_2016} was proposed as an extension to ShapeNet~\cite{chang_shapenet_2015} with part-level annotations, but only contains coarse-grained part segmentations extracted using a deep active learning framework.
  In contrast, PartNet~\cite{mo_partnet_2018} builds on ShapeNet~\cite{chang_shapenet_2015} and provides fine-grained part segmentation labels, but similarly does not contain material information.

  \noindent Material information offers several distinct advantages.
  First, it provides extra semantic information about an object, which enables a variety of important 3D object understanding tasks.
  Second, it helps create more realistic renderings, making the models better suited for transferring from synthetic to real scenarios.
  Finally, applying different materials to the same geometric 3D shape can be treated as a special form of training data augmentation.
  Current datasets lack part-level material information which underscores the need for a new resource.
  Our dataset fills this gap and invites researchers to explore new challenges and opportunities in 3D visual understanding.
  \compat is an extension of the 3DCoMPaT~\cite{3dcompat_2022} dataset, which was previously published at a conference.\\[-1em]

\begin{figure*}[h!]
    \centering
    \includegraphics[width=\textwidth]{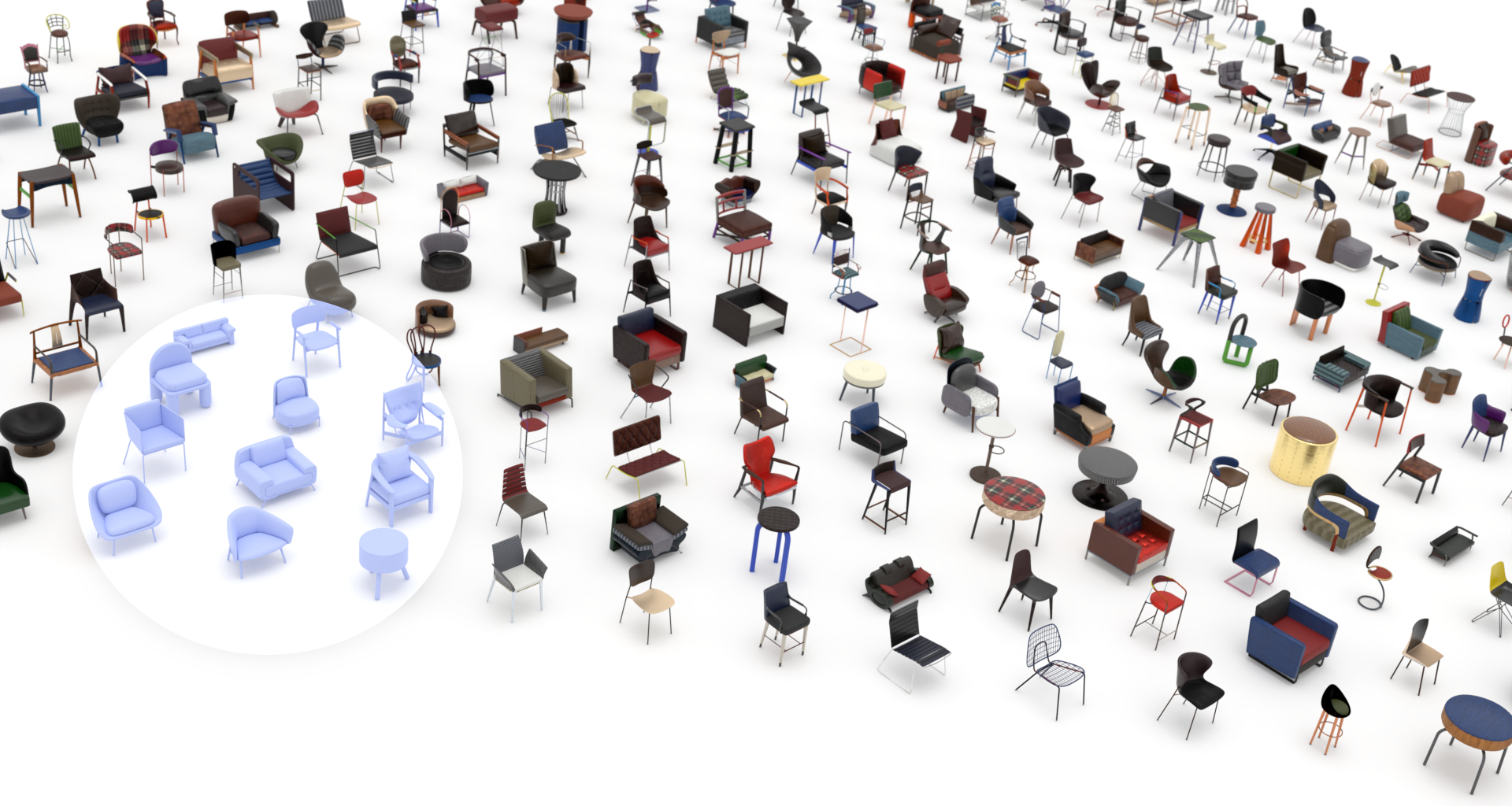}
    \vspace{-1em}
    \caption{\textbf{Rendering of randomly sampled shapes from \compat.}
    The dataset comprises a rich collection of stylized 3D shapes annotated at the part-instance level.
    These renderings demonstrate the varying shapes, styles, and materials that are captured, enabling comprehensive exploration and analysis of compositional 3D vision tasks. Shapes are consistently aligned across classes and orientations are consistent for all 3D models.
    In the left circle, we illustrate the untextured 3D geometries we start from as a reference.
    We provide additional reference shapes from all \nshapeclasses shape categories in Figure \ref{fig:all_classes} of the appendix.
    }
    \label{fig:renderings}
  \end{figure*}

\begin{figure*}
    \centering
    \includegraphics[width=\textwidth]{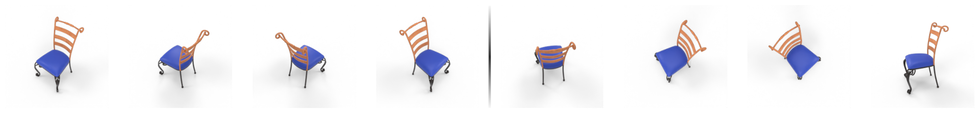}
    \vspace{0em}
    \caption{\textbf{Canonical and random views.}
    Canonical (\textbf{left}) and random (\textbf{right}) views rendered for a single stylized 3D shape.
    Random viewpoints are sampled, while canonical viewpoints are equally spaced around the shape.
    In Figure \ref{fig:views_additional} of the appendix, we provide additional examples of canonical and random views.
    }
    \label{fig:views}
    \vspace{-1em}
  \end{figure*}

  \noindent We introduce a new richly annotated multimodal 2D/3D large-scale dataset: \textbf{\compat}, standing for \textbf{Co}mpositions of \textbf{M}aterials on \textbf{Pa}rts of \textbf{3D} \textbf{T}hings.
  Our dataset comprises \nstylizedshapes stylized 3D shapes rendered from \nviewpermodel views, across \nshapeclasses shape categories, \npartsfine unique fine-grained part names and \npartscoarse coarse-grained part names, and \nmaterialclasses unique materials from \nmaterialclassescoarse material classes.
  We sample object-compatible combinations of part-material pairs to create \avgncomposition styles per shape.
  Each object with the applied part-material composition is rendered from \nviewpertype equally spaced views and \nviewpertype random views.
  We render images for a \avgncomposition total compositions, leading to \nviews\footnote{Figure detail: $\nshapes\ (\# \texttt{shapes}) \times \avgncomposition\ (\# \texttt{styles}) \times \nviewpermodel\ (\# \texttt{views}) \\ \times 2\ (\# \texttt{part semantic levels}) = \nviewsshort\ \text{views}$} total rendered views.
  Examples of some rendered compositions and views are provided in Figures \ref{fig:renderings} and \ref{fig:views} respectively.\\

  \point{Dataset.} To create our dataset, we start with \nshapesshort unique geometries which we segment at a fine-grained part level into a total of \npartsfine segmented parts (leading to \npartinstances average part instances per shape).
  For each part of each shape, human experts determine a list of compatible/applicable materials.
  Then, we generate a stylized model by sampling over the compatible materials of each part with a limit of \avgncomposition styles per shape, leading to \nstylizedshapes stylized shapes.\\

  \point{Previous work.} Our proposed dataset differs from previous work in numerous ways.
  Our dataset contains a diverse set of high-quality materials: for each part found in every 3D shape, we annotate possible compatible materials that may be applied to each part, allowing us to generate multiple material combinations for a single shape (we refer to a combination of materials (\textit{composition}) applied to a model as a \textit{style}).
  We also enrich our dataset with 2D renders, depth maps, part masks, and material masks for each rendered view, and hierarchical part and material annotations in both the 2D and 3D modalities.

  \noindent In summary, our \compat dataset can be distinguished from existing datasets by the following four key aspects:

  \noindent \semipoint{a} \textit{Human-generated vs. 3D scanned geometry.} ScanNet~\cite{dai_scannet_2017} and Matterport3D~\cite{chang_matterport3d_2017} datasets are scanned 3D geometry datasets. Conversely, ShapeNet~\cite{chang_shapenet_2015} and our \compat dataset are human-created, mostly by professional 3D modelers. Human-created geometry is generally of higher quality and has fewer artifacts, but is however more expensive and time-consuming to collect.
  For the Objaverse~\cite{deitke_objaverse_2022,deitke_objaverse-xl_2023} dataset, the authors thus propose to scrape 3D models from well-known web repositories which are mostly created by artists.
  However, the quality of these models is not guaranteed, and models are not annotated with part-level information.
  The realism of the collected objects is also not a given, as models in these repositories are not designed to be realistic, but rather to be visually appealing as they are typically targeted at the video game industry.
  Our dataset is human-generated and is designed to be realistic, and comprises high-quality textures and geometry.

  \noindent \semipoint{b} \textit{Part segmentation information.}
  For some datasets, none or only a subset of the shapes have segmented part information, which is an important feature of datasets like PartNet~\cite{mo_partnet_2018} and is also a core characteristic of our dataset.
  We provide part segmentation information following two semantic levels, in both 2D and 3D modalities.

  \noindent \semipoint{c} \textit{Texture coordinates, textures, and materials.}
  A key focus of our work lies in the stylization of 3D shapes with appropriate texture coordinates, textures, and materials.
To achieve a superior level of quality when rendering numerous material compositions on each shape, our models are equipped with human-verified texture coordinates and part-wise material compatibility information.
  While previous attempts have been made to enhance a subset of ShapeNet with part-wise material information~\cite{lin_partmat_2018}, it falls short in comparison to our work in terms of the number of shapes (3080 vs. \nshapes), shape classes (3 vs \nshapeclasses), and materials (6 vs. \nmaterialclassescoarse, and \nmaterialclasses fine-grained annotated classes).

  \noindent \semipoint{d} \textit{Automatically generated vs. human-generated annotations.} \compat shapes are annotated manually by a team of trained humans.
  Part names are consistent across and within categories, and are defined in shape category-specific guidelines.
  Each guideline is defined by a team of researchers and professional modelers, and contains rigorous definitions and examples for each part that may occur within a given shape category.
  All models are manually segmented at a part level rather than with deep learning models like OpenRooms~\cite{li_openrooms_2021} or ShapeNet-Part~\cite{yi_shapenet_part_2016}.\\ 

  \point{Grounded CoMPaT Recognition.} We introduce a novel task called CoMPaT recognition, which focuses on collectively recognizing and grounding shape categories along with the associated part-material pairs composing them.
  In Figure \ref{fig:gcr_task}, we illustrate the task with an example.
  Given an input shape, the task aims to recognize both the shape category and all part-material pairs composing it. In the example shown, an agent first needs to identify the shape as a chair, and then all part-material pairs, such as a "seat" made of "leather" and a "backrest" made of "fabric".
  This novel task, compatible with both 2D and 3D modalities, goes beyond recognition with a grounded variant requiring the precise segmentation of parts alongside the recognition of associated materials.\\

  \point{Contributions.} Our work introduces a new dataset, and introduces the GCR recognition task. The contributions of this work can be summarized in the following points:\\
  \vspace{-0.75em}

  \begin{itemize}
    \vspace*{-0.5em}
    \setlength\itemsep{1.2em}
    \item We propose a new dataset comprised of \nstylizedshapes stylized models to study the composition of materials on parts of 3D objects. 
    Our dataset contains
    (a) a diverse set of \nmaterialclasses\ materials for 3D shapes, where (b) material assignment is done at a coarse and fine-grained part-level;
    (c) segmentation masks in 2D and 3D, alongside (d) human-verified texture coordinates.
    \item We validate our dataset with experiments covering 2D and 3D vision tasks, including object classification, part recognition and segmentation, material tagging and shape generation.
    \item We also propose \textbf{G}rounded \textbf{C}oMPaT \textbf{R}ecognition (GCR), a novel task aiming at collectively recognizing and grounding compositions of materials on parts of 3D objects.
    We introduce two variants of this task, and leverage 2D/3D state-of-the-art methods as baselines for this problem. 
  \end{itemize}

\section{Related work}
  \begin{table*}[h]
    \centering
    \caption{
      \vspace{0.5em}
      \textbf{Comparison of \compat with existing 3D datasets.}
      \textit{Multi-level} indicates whether the dataset contains hierarchical part semantics. \textit{Alignment} indicates whether the dataset contains aligned 2D/3D data.
      We denote by \noanno missing annotations, \textit{e.g.} cases where the dataset contains textured shapes, but does not include material annotations.
      Among datasets with part-level annotations, \compat contains the largest number of stylized shapes, object categories, materials and also provides a large collection of aligned 2D images.
    }
    \setlength{\extrarowheight}{0.3em}
    \resizebox{\linewidth}{!}{
      \begin{tabular}{@{}l  r  c  r  c  l  c  c  l  c  c  c l  r  c @{}}
        \toprule
        \multicolumn{1}{c}{
            \multirow{2}{*}{\textbf{Dataset}}}
        &  \multicolumn{4}{c}{{\color{IndianRed}{ \faCube }}\ \textbf{Shapes}}
        &  & \multicolumn{2}{c}{{\color{DarkOrange}{ \faPalette }}\ \textbf{Materials}}
        &  & \multicolumn{3}{c}{{\color{LightSteelBlue}{ \faPuzzlePiece }}\ \textbf{Parts}}
        &  & \multicolumn{2}{c}{{\color{DarkOrchid}{ \faFileImage }}\ \textbf{Images}}
        \\ \cmidrule(lr){2-5} \cmidrule(lr){7-8} \cmidrule(lr){10-12} \cmidrule(l){14-15} 

        \multicolumn{1}{c}{} & \multicolumn{1}{c}{\textbf{Count}}
        & \multicolumn{1}{c}{\textbf{Stylized}}
        & \multicolumn{1}{c}{\textbf{Classes}}
        & \textbf{Source}
        
        &  & \multicolumn{1}{c}{\textbf{Count}} & \multicolumn{1}{c}{\textbf{Classes}}
        &  & \multicolumn{1}{c}{\textbf{\#Instances/Shape}} & \multicolumn{1}{c}{\textbf{Multi-level}} & \multicolumn{1}{c}{\textbf{Instances}}
        &  & \multicolumn{1}{c}{\textbf{Count}} & \textbf{Alignment} \\ \midrule

        ModelNet~\cite{wu_modelnet_2015}                           & 128K    & \nodata & 662     & modelled  &  & \nodata & \nodata  &  & \nodata & \nodata & \nodata & & \nodata & \nodata \\
        ShapeNet-Core~\cite{chang_shapenet_2015}                   & 51,3K   & \nodata & 55      & modelled  &  & \noanno & \noanno  &  & \nodata & \nodata & \nodata & & \nodata & \nodata \\
        ShapeNet-Sem~\cite{chang_shapenet_2015}                    & 12K     & \nodata & 270     & modelled  &  & \noanno & \noanno  &  & \nodata & \nodata & \nodata & & \nodata & \nodata \\
        PhotoShape~\cite{park_photoshape_2018}                     & 5,8K    & 29K     & 1       & modelled  &  & 658     & 8        &  & \nodata & \nodata & \nodata & & \nodata & \nodata \\[0.2em]
        \midrule               
                    
        GSO~\cite{downs_gso_2022}                                  & 1K      & \nodata & 17   & scanned  &  & \noanno & \noanno      &  & \nodata & \nodata & \nodata & & \nodata & \nodata \\
        OmniObject3D~\cite{wu_omniobject3d_2023}                   & 6K      & \nodata & 190  & scanned  &  & \noanno & \noanno      &  & \nodata & \nodata & \nodata & & \nodata & \nodata \\[0.2em]
        \midrule               
                    
        ObjectNet3D~\cite{leibe_objectnet3d_2016}                  & 44,2K   & \nodata & 100     & modelled  &  & \nodata  & \nodata &  & \nodata & \nodata & \nodata & & 90K  & pseudo \\
        3D-Future~\cite{fu_3dfuture_2020}                          & 9,9K    & \nodata & 15      & modelled  &  & \noanno  & 15      &  & \nodata & \nodata & \nodata & & 20K  & exact  \\
        ABO~\cite{collins_abo_2022} + HAL3D~\cite{yu_hal3D_2023}   & 8K      & \nodata & 63      & modelled  &  & \noanno  & \noanno &  & \noanno & \noanno & \noanno & & 398K & pseudo \\
        Objaverse-XL~\cite{deitke_objaverse-xl_2023}               & 10,2M   & \nodata & \noanno & modelled  &  & \noanno  & \noanno &  & \nodata & \nodata & \nodata & & 66M  & exact  \\[0.2em]
        \midrule

        ShapeNet-Part~\cite{yi_shapenet_part_2016}                & 31,9K   & \nodata & 16  & modelled  &  & \nodata & \nodata      &  & 2.99    & \nodata & \nodata & & \nodata & \nodata \\
        ShapeNet-Mats~\cite{lin_partmat_2018}                     & 3,2K    & \nodata & 3   & modelled  &  & \noanno & 6            &  & 6.20     & \nodata & \nodata & & \nodata & \nodata \\
        PartNet~\cite{mo_partnet_2018}                            & 26,7K   & \nodata & 24  & modelled  &  & \nodata & \nodata      &  & 7.21    & \icoyes & \icoyes & & \nodata & \nodata \\[0.2em]
        % 3DCoMPaT \cite{3dcompat_2022} & 7,2K & 7,2M & 43 & modelled &  & 167 & 11 &  & 5.12 & \nodata & \nodata & &  58M & exact  \\
    
        \textbf{\compat} 
        & \nshapesshort & \textbf{\nstylizedshapesshort} & \textbf{\nshapeclasses} & modelled 
        &  & \textbf{\nmaterialclasses} & \textbf{\nmaterialclassescoarse}
        &  & \textbf{\npartinstances} & \icoyes
        & \icoyes & &  \textbf{\nviewsshort} & exact  \\[0.2em] \arrayrulecolor{black}\bottomrule
      \end{tabular}
    }
    \vspace{0em}
    \label{tab:dataset_comparison}
  \end{table*}

  \begin{figure*}
    \vspace{-1em}
    \centering
    \subfloat{%
      \hspace{-0.4em}
      \includegraphics[width=0.24\linewidth]{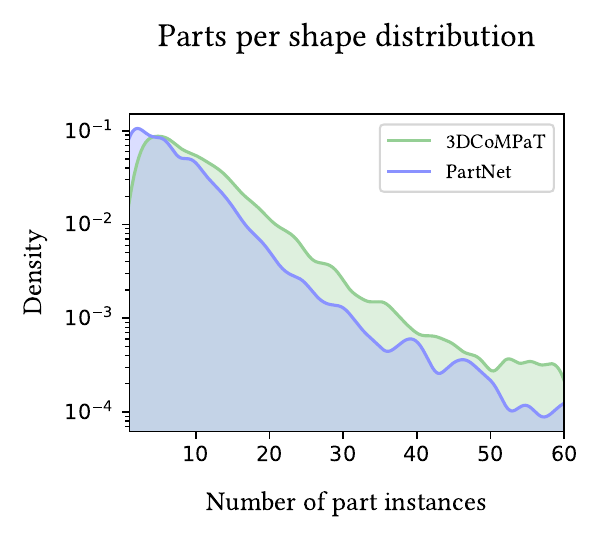}
      \hspace{0.4em}
      {\color{mygrey}\rule[10pt]{0.5pt}{100pt}}
    }
    \subfloat{%
      \includegraphics[width=0.24\linewidth]{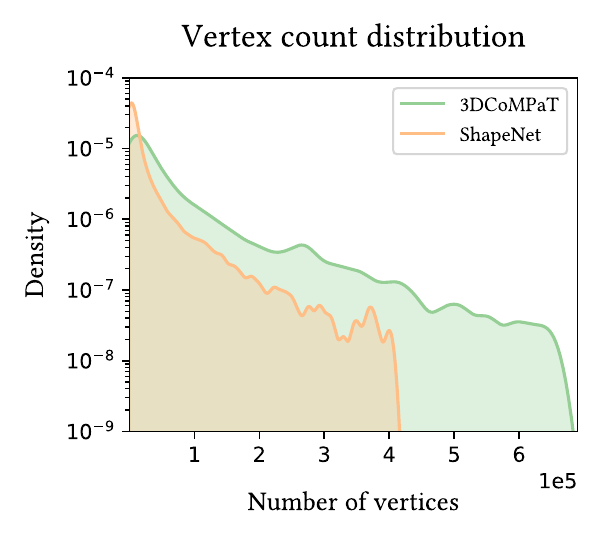}
    }
    \subfloat{%
      \includegraphics[width=0.24\linewidth]{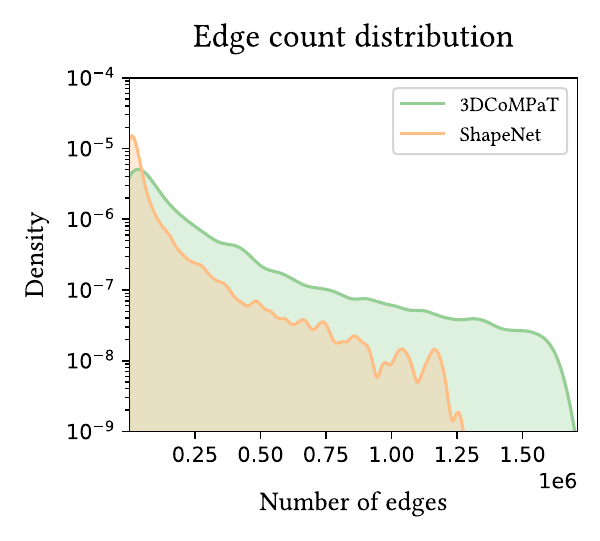}
    }
    \subfloat{%
      \includegraphics[width=0.24\linewidth]{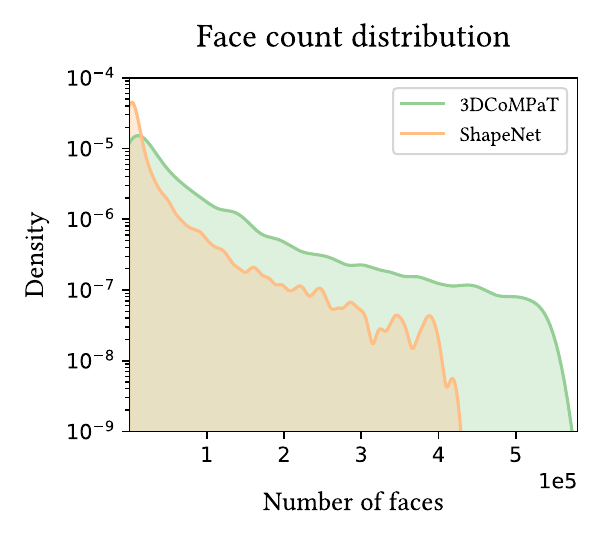}
    }
    \vspace{-1.25em}
    \caption{\textbf{Comparison with PartNet.} Part instances per shape distributions compared to PartNet~\cite{mo_partnet_2018} (\textbf{left}), Density plots depicting the distribution of vertex counts, edge counts, and face counts across 3D shapes extracted from both the \compat and ShapeNet datasets (\textbf{right}). We show significantly higher numbers of vertices, edges, and faces in \compat shapes compared to ShapeNet. All annotated shapes in PartNet originate from ShapeNet.} 
    \label{fig:poly_counts}
  \end{figure*}

  \fpoint{Early efforts.} Several datasets have been initially proposed to facilitate 3D visual understanding, such as ShapeNet~\cite{chang_shapenet_2015}, ModelNet~\cite{wu_modelnet_2015}, and PartNet~\cite{mo_partnet_2018}.
  ModelNet~\cite{wu_modelnet_2015} is one of the first datasets of 3D objects, and includes 40 shape categories and 12K unique 3D shapes.
  ShapeNet~\cite{chang_shapenet_2015} is a large-scale dataset of 3D textured objects, with 55 shape categories and 51K unique 3D shapes. ShapeNet is annotated at the shape-level, and categories are extracted from WordNet~\cite{miller_wordnet_1995}.
  It has emerged as an important benchmark for deep learning-based modeling, representation, and generation of 3D shapes.
  ObjectNet3D~\cite{leibe_objectnet3d_2016} is an object-centric dataset of 3D CAD models with 100 shape categories and 90K unique 3D shapes, and approximate 2D-3D image alignments.
  ModelNet, ShapeNet and ObjectNet3D are object-centric datasets, and do not contain part-level annotations.\\

  \point{Part-understanding.} In an attempt to bridge this gap, ShapeNet-Part~\cite{yi_shapenet_part_2016} was first proposed as an extension to ShapeNet~\cite{chang_shapenet_2015} with part-level annotations.
  It contains 16 shape categories and 31K 3D shapes, but part annotations are only provided at a coarse-grained semantic level, and are extracted using a deep active learning framework instead of human annotation.
  PhotoShape~\cite{park_photoshape_2018} is one of the earliest efforts in gathering 3D shapes with high-quality textures.
  It contains 5.8K 3D shapes from 29 shape categories, and proposes to transfer material properties regressed from real images to untextured 3D shapes.
  PartNet~\cite{mo_partnet_2018} was built as a large-scale dataset of 3D shapes annotated with fine-grained, instance-level, and hierarchical part segmentations.
  PartNet is also created on top of ShapeNet~\cite{chang_shapenet_2015} and contains 26K 3D shapes from 24 shape categories. PartNet is a valuable resource for advancing research in 3D shape analysis and understanding.
  Our work stands apart from PartNet in three main ways:
  \vspace{0.5em}
  \begin{itemize}
    \setlength\itemsep{0.5em}
    \item \semipoint{a} We provide both coarse-grained and fine-grained material information for each part of each shape.
    \item \semipoint{b} We enrich 3D shapes with 2D renders, part masks, material masks, and depth maps.
    \item \semipoint{c} We use a human verification process to ensure the compatibility of sampled materials with each part of each shape.
    \vspace{0.5em}
  \end{itemize}
  
  \point{High-resolution datasets.} In an effort to provide high-quality and realistic shapes and textures, OmniObject3D~\cite{wu_omniobject3d_2023} and ABO~\cite{collins_abo_2022} datasets introduced 3D assets with rich, high-quality textures. ABO~\cite{collins_abo_2022} also provides unlabelled part instance segments for a subset of 3,4K shapes, in the form of unnamed connected shape pieces. HAL3D~\cite{yu_hal3D_2023} builds on ABO using an active learning pipeline to provide semantic part instance names for ABO, but did not release the dataset to the public.
  Google Scanned Objects~\cite{downs_gso_2022} is a scanned dataset of reconstructed 3D objects with high-quality textures and geometries, and contains 1K 3D shapes from 17 diverse categories of small objects.
  OmniObject3D~\cite{wu_omniobject3d_2023} is a scanned dataset of 3D objects with high-quality textures, and contains 6K 3D shapes from 190 shape categories based on ImageNet~\cite{deng_imagenet_2009} and LVIS~\cite{gupta_lvis_2019}.
  ABO~\cite{collins_abo_2022} is a dataset of 3D objects with high-quality textures and geometries, and contains 8K 3D shapes from 63 shape categories based on product catalogs extracted from Amazon.

  \noindent 3D-Future~\cite{fu_3dfuture_2020} presented a dataset comprising 10K industrial 3D CAD shapes of furniture developed by professional designers. More recently, Objaverse~\cite{deitke_objaverse_2022} and Objaverse-XL~\cite{deitke_objaverse-xl_2023} expanded the horizon of 3D object datasets by releasing over 10 million artist-designed 3D objects with high-quality textures.

  \noindent Despite these significant strides in advancing 3D understanding, these modern object-centric 3D datasets (e.g.~\cite{chang_shapenet_2015, wu_modelnet_2015, deitke_objaverse_2022, deitke_objaverse-xl_2023}) and scanned datasets (e.g.~\cite{dai_scannet_2017},~\cite{chang_matterport3d_2017},~\cite{wu_omniobject3d_2023}) lack part-level annotations.
  PartNet~\cite{mo_partnet_2018}, building on ShapeNet~\cite{chang_shapenet_2015}, offers fine-grained part segmentations of 3D meshes but does not include material information.
  The absence of such part-level annotations and material data points to the significance of a dataset like \compat, which bridges these gaps and serves as a comprehensive resource for furthering research in 3D visual understanding.\\ 

  \begin{figure*}
    \centering
    \hspace{0.75em}
    \includegraphics[width=0.95\linewidth]{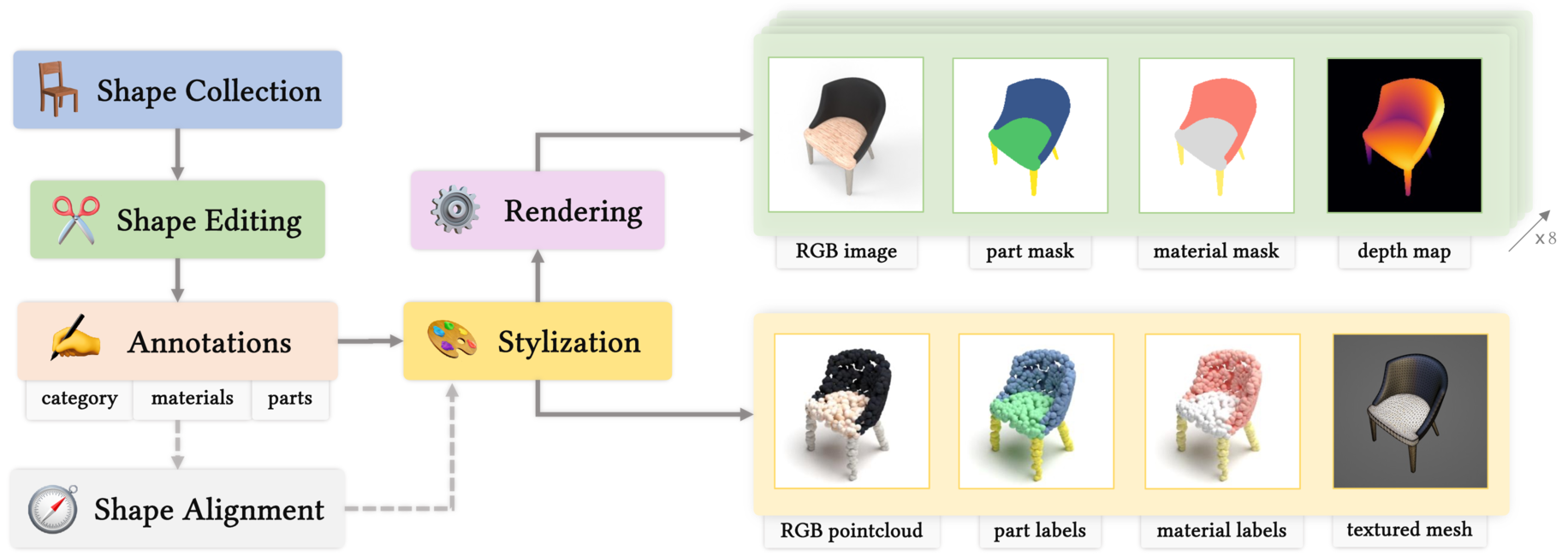}
    \caption{\textbf{Detailing the data pipeline of \compat.}
    Starting from the collection of 3D shapes, we perform a first editing step consisting of model re-scaling, UV map correction and removal of undesirable meshes.
    Material compatibility information is also collected for each part of each shape, alongside the shape category.
    Shapes are then annotated at a fine-grained part-instance level, and part names are iteratively refined and uniformized using a web-based shape visualizer.
    Misaligned shapes are semi-automatically realigned using part annotations as a prior.
    Finally, we sample a set of materials for each part of each shape, and render each stylized shape from multiple viewpoints.
    }
    \label{fig:data_pipeline}
    \vspace{-1em}
  \end{figure*}

  \point{Comparison with existing work.} In Table \ref{tab:dataset_comparison}, we compare \compat with existing prevalent 3D datasets.
  We distinguish between datasets originating from 3D artists (first group), scanned objects datasets (second group), datasets with aligned 2D images (third group), and datasets with part-level annotations (fourth group).
  We scrutinize fundamental aspects, including the number of shapes provided, whether or not stylized shapes are included, the number of classes represented and whether shapes come from scans or are designed from CAD tools.
  Additionally, we assess the availability of material annotations. We differentiate cases where textured shapes are provided but without material annotations (\noanno) like in GSO~\cite{downs_gso_2022} and Objaverse-XL~\cite{deitke_objaverse-xl_2023}, from cases where they are provided at a coarse-grained level only (e.g. 3D-Future~\cite{fu_3dfuture_2020} in which only coarse material annotations are available), or are provided at both coarse and fine-grained levels.
  Material annotations and part-wise material annotations are important as they provide essential contextual information about the surface properties and appearance of objects, facilitating compositional understanding and analysis of 3D shapes.

  \noindent We also consider the inclusion of aligned 2D images, and differentiate between cases where images are pseudo-aligned or exactly aligned with matching 3D shapes.
  Pseudo-alignment includes using a manual 3D alignment pipeline with close candidate CAD models~\cite{leibe_objectnet3d_2016}, or using an automatic 3D alignment strategy with exactly matching shapes (e.g. based on differentiable rendering~\cite{collins_abo_2022}).
  Exact alignments are achieved by producing synthetic 2D images from 3D models using a rendering engine, and then projecting the 3D models into the 2D images using the camera pose ground truth (e.g. 3D-Future~\cite{fu_3dfuture_2020}, this work). In contrast to other works, \compat emerges distinctively by offering a large collection of \nshapesshort stylized shapes, each accompanied by complete multi-level part-material information.
  With PartNet, \compat is the only dataset with instance-level part annotations, which are essential in tasks involving denumerating parts composing a shape.
  Notably, \compat also offers a large collection of aligned 2D/3D data with \textbf{\nviewsshort images} and \textbf{\nstylizedshapesshort} shapes, enabling its use in diverse multi-modal learning applications benefiting from scale like object classification, part recognition or novel view synthesis.\\
  
  \point{Mesh resolutions.} In Figure \ref{fig:poly_counts}, we compare part instances per shape distributions between \compat and PartNet~\cite{mo_partnet_2018} (left), and model resolution statistics for 3D CAD models from \compat and ShapeNet~\cite{chang_shapenet_2015} (right).
  This comparison is important because ShapeNet~\cite{chang_shapenet_2015} serves as the CAD model data source for PartNet~\cite{mo_partnet_2018} and ShapeNet-Part~\cite{yi_shapenet_part_2016}, which are two of the most prominent datasets for 3D part understanding.
  We provide density plots over vertex counts, edges counts, and faces counts for 3D CAD models from both datasets.
  We show that \compat exhibits both higher numbers of part instances per shape compared to PartNet and significantly higher average numbers of vertices, edges, and faces when compared to ShapeNet.
  While polygon count is not a perfect proxy for shape visual quality and realism, it is a useful metric for comparing the relative complexity of meshes in each dataset.
  This quantitative assessment underscores the richness of annotations and geometries within \compat, making it a valuable resource for advancing research in 3D shape analysis and understanding.
  
\section{\compat}
\label{sec:dataset}

The \compat dataset is based on a collection of artist-designed 3D CAD models collected and annotated
in collaboration with an industry partner.
It contains \nshapesshort geometries annotated and segmented at a fine-grained part-instance level,
with material compatibility information for each annotated part.
For each shape, \nviewpermodel rendered views are provided from canonical and random viewpoints (see Figure \ref{fig:views}).
For each rendered view, depth maps, part maps, and material maps are rendered (see Figures \ref{fig:2D_data_example} and \ref{fig:full_data_example_appendix}).

\noindent All annotations are provided by trained annotators following a rigorous multi-stage review process.
\compat is a richly annotated, multimodal 2D/3D dataset: In Figure \ref{fig:all_modalities}, we illustrate all data provided for a single stylized shape from our dataset.

\subsection{Dataset}

\begin{figure} 
  \vspace{-2em}
  \centering
  \includegraphics[width=\linewidth]{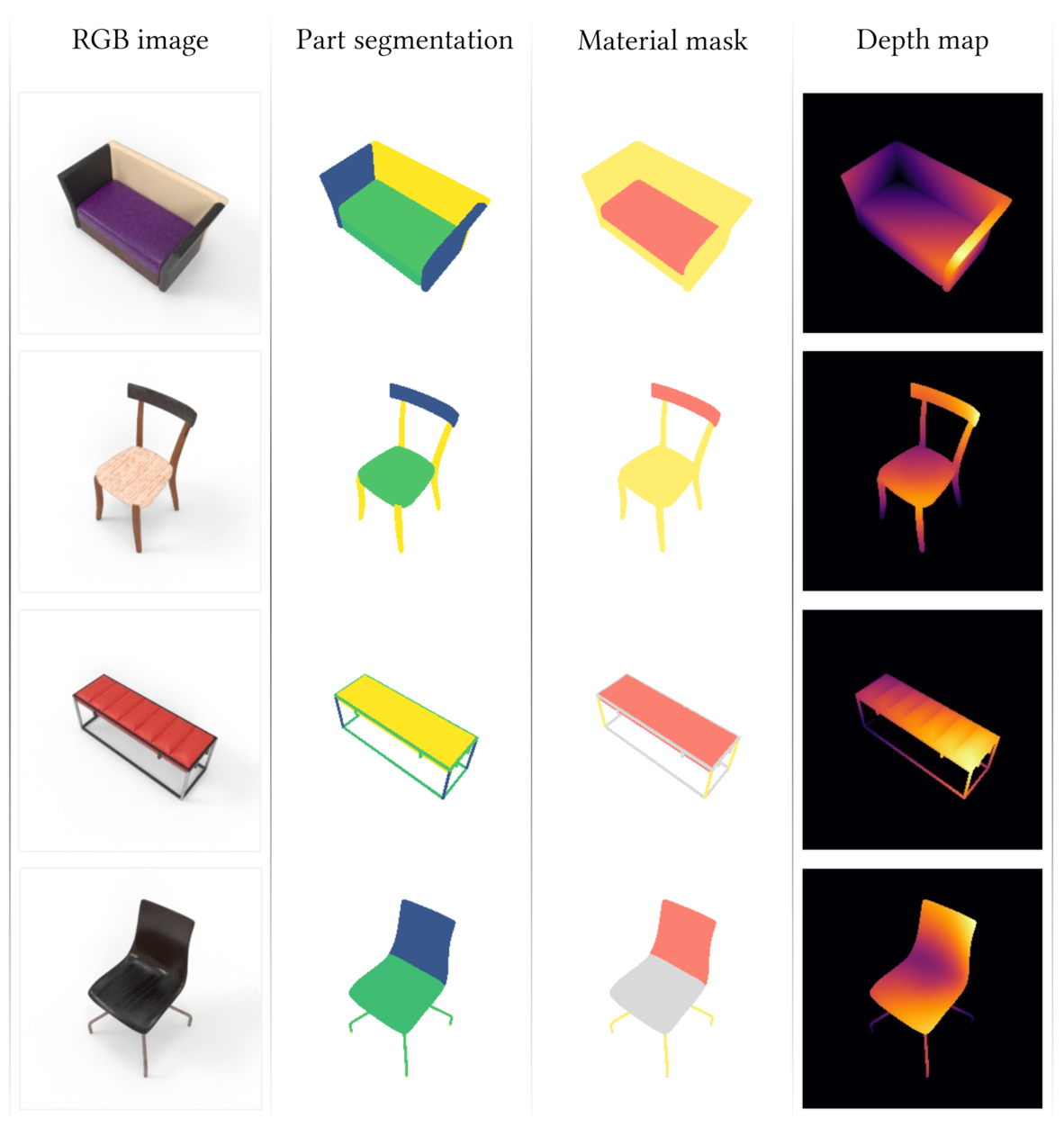}
  \vspace{-2em}
  \caption{\textbf{2D data associated with each viewpoint.} Each 2D image is complemented by corresponding part segmentation masks, material masks, and absolute depth maps, enabling various 2D/3D vision tasks. Camera parameters are also provided for each render. See Figure \ref{fig:full_data_example_appendix} in the appendix for additional examples.}
  \label{fig:2D_data_example}
  \vspace{-1em}
\end{figure}

\begin{figure}
  \centering
  \vspace{0.5em}
  \includesvg[width=0.99\linewidth]{assets/figs_svg/part_count_comparison.svg}
  \hspace{-1em}
  \includesvg[width=0.99\linewidth]{assets/figs_svg/part_occurrences.svg}
  \vspace{-2em}
  \caption{
    \textbf{Distribution of part occurrences for fine/coarse levels.}
    We plot the average number of unique parts per object for fine and coarse semantic levels, across all shape categories (\textbf{top}).
    We also plot the sorted number of occurrences of each part (logscale, \textbf{bottom}).
  }
  \label{fig:part_occ}
  \vspace{0em}
\end{figure}

\begin{figure}
  \vspace{-1em}
  \centering
  \includesvg[width=0.99\linewidth]{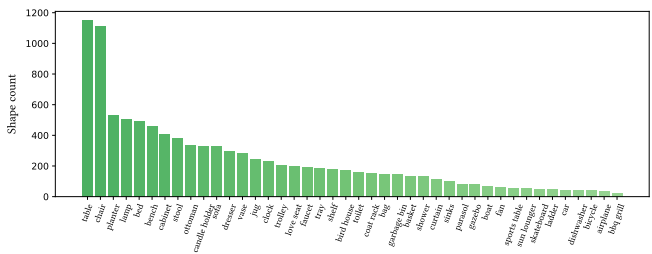}
  \vspace{-2em}
  \caption{
    \textbf{Distribution of shape category occurrences.}
    We plot the sorted number of occurrences of each shape category.
    The class distribution in \compat is significantly long-tailed.
  }
  \label{fig:class_occ}
  \vspace{-1em}
\end{figure}

\fpoint{3D Data.}
Alongside each stylized shape, we provide a part-segmented textured 3D mesh, an RGB pointcloud, and point-wise and triangle-wise part and material annotations.
All part segmentation information is provided in \textit{coarse-grained} and \textit{fine-grained} semantic levels.
RGB pointclouds can be resampled at any resolution starting from the available textured 3D meshes.
In Figure \ref{fig:all_modalities}, we illustrate the 3D data provided for a single stylized shape.\\

\point{2D Data.}
Each stylized shape is rendered from \nviewpermodel viewpoints: \nviewpertype canonical viewpoints and \nviewpertype random viewpoints.
Canonical viewpoints are equally spaced around the shape.
Random viewpoints are sampled uniformly on the upper hemisphere centered on the center of the shape's bounding box.
In Figure \ref{fig:2D_data_example}, we showcase the 2D data provided for the first \textit{canonical} viewpoint across four different 3D shapes.
Each 2D image is accompanied by part segmentation masks, material masks, and depth maps.
For each image, camera parameters are also provided.
Part segmentation masks and material masks are available in two semantic levels: \textit{coarse-grained} and \textit{fine-grained}.\\

\vspace{-1.5em}
\subsection{Data collection pipeline}

\begin{figure}
  \vspace{-0.5em}
  \centering
  \includegraphics[width=\linewidth]{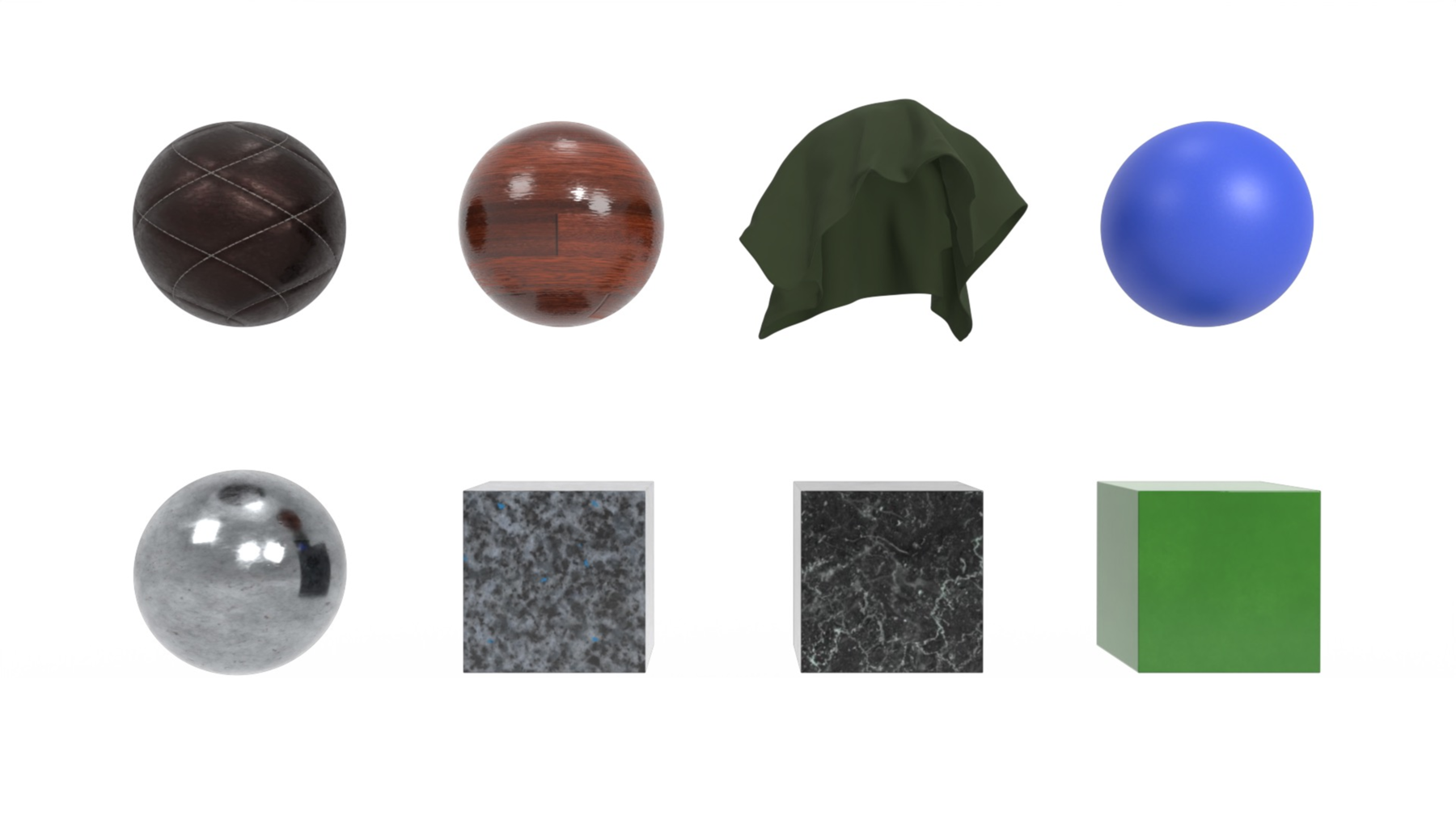}
  \vspace{-1.25em}
  \caption{
    \textbf{Example coarse materials in \compat.}
    We randomly sample a single material from  8 of the \nmaterialclassescoarse materials provided in \compat, and render it on demo primitives.
    From left to right, top to bottom:\ \texttt{leather, wood, fabric, plastic, metal, granite, marble, ceramic}.
  }
  \label{fig:materials_render}
  \vspace{-1em}
\end{figure}

The complete data collection pipeline is depicted in Figure \ref{fig:data_pipeline}, and includes the following steps:
\begin{itemize}
  \setlength\itemsep{0.2em}
  \vspace{0.5em}
  \item \textbf{Collection and Editing.} 3D shapes are collected and edited by our industry partner.
  \item \textbf{Part annotations.} Annotators follow each category-level guideline when adding instance-level part annotations and segmentations to each shape.
  \item \textbf{Material assignments.} Annotators select compatible materials for each part of each shape, from among \nmaterialclassescoarse possible coarse classes.
  \item \textbf{Stylized shapes.} We sample a set of fine-grained materials for each part of each shape, which we refer to as a \textit{style}.
  \item \textbf{Rendering.} We render each shape from multiple viewpoints with matching masks, depth maps and pointcloud data, as detailed in Section \ref{subsec:rendering}.
\end{itemize}

\begin{figure*}
  \vspace{0em}
  \hspace{-0.25em}
  \centering
  \hspace{-0.3em}
  \includegraphics[width=\linewidth]{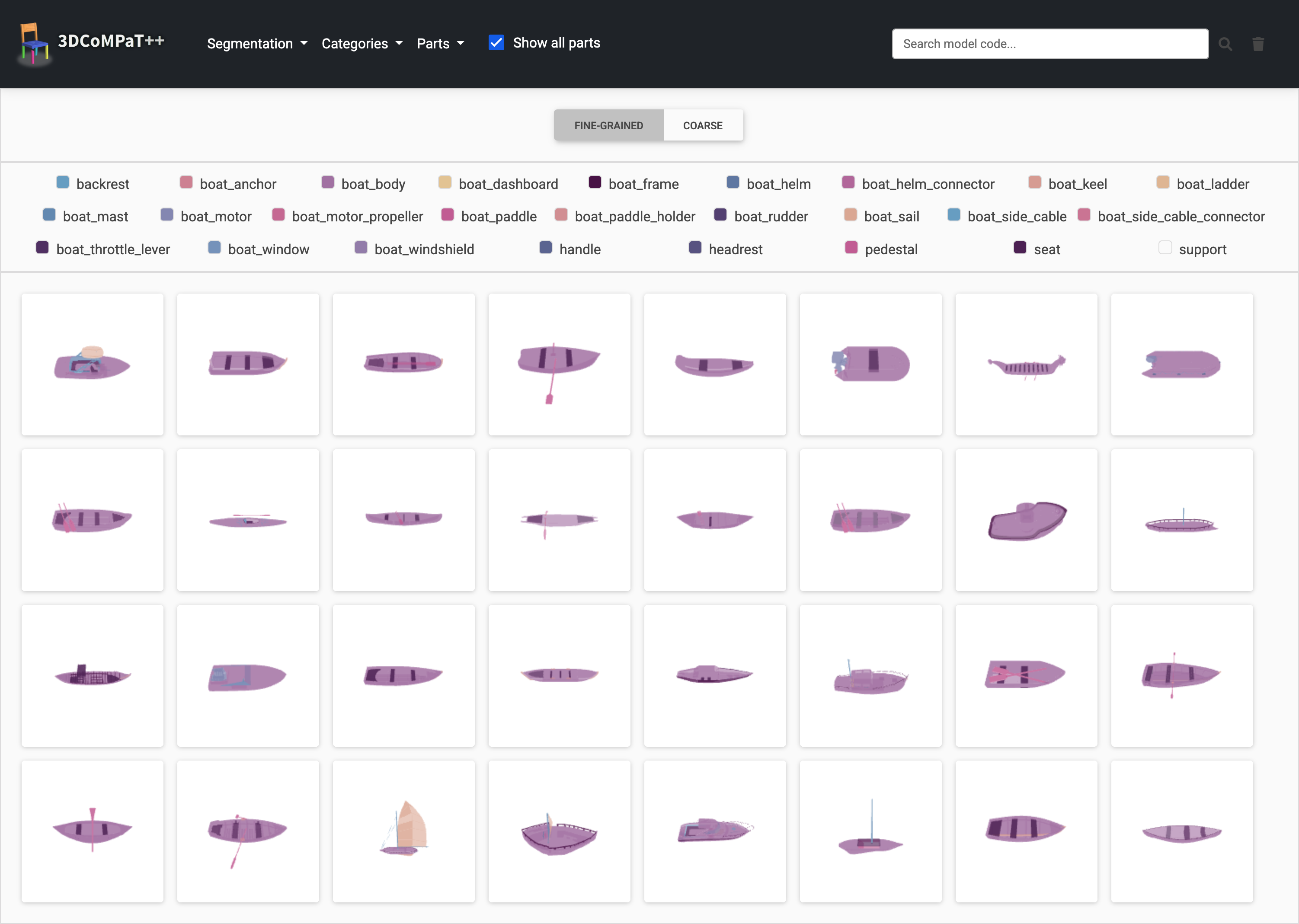}
  \vspace{-1em}
  \caption{
    \textbf{Interactive web-based browser for \compat.}
    We provide an interactive 3D shape browser built using Three.JS~\cite{threejs_2023} to visualize the 3D shapes in our dataset.
    The browser allows to easily visualize instance and semantic part annotations in both coarse and fine-grained levels.
  }
  \label{fig:web_browser}
  \vspace{-1em}
\end{figure*}

\vspace{1em}
\point{Collection and Editing.}
All 3D shapes are collected by our industry partner.
Editing steps include model scaling,
the correction of UV maps,
the removal of undesirable/invalid meshes in the shape
(e.g., additional objects like a vase on top of a table), etc.
Furthermore, as visible in Figure \ref{fig:renderings}, all shapes are consistently aligned across classes and orientations are consistent for all 3D models. 
To align shapes, we use part annotations as a prior to automatically rotate a majority of misaligned shapes (for example, using the fact that the "\texttt{vertical\_back\_panel}" part should appear at the back of a shape).
We then manually adjust the remaining misaligned shapes by using a web visualization tool\footnote{The 3DCoMPaT annotated shapes web-based browser is accessible here: \url{https://3dcompat-dataset.org/browser/}} (see Figure \ref{fig:web_browser}).
3D shapes are also scaled to fit within a unit cube centered at the world origin.\\

\point{Part annotations.}
By combining expert knowledge with the analysis of unannotated shapes,
we define fine-grained part-level guidelines.
A guideline is defined for each shape category and provides a non-ambiguous definition
of each possible fine-grained part that can occur in shapes belonging to the category.
Annotators follow each category-defined guideline when adding instance-level part annotations
and segmentations to each shape (see Figure \ref{fig:guideline_short} for an example of a shape guideline for the \texttt{faucet, shower, sink} shape categories).
Part segments and names are iteratively refined using a web-based shape visualizer (see Figure \ref{fig:web_browser}).
This browser allows reviewers to visualize segments for a specific part class in a shape category,
allowing to efficiently verify part semantics consistency across shapes and quickly identify
annotation errors.
Corner cases, when frequent enough, are identified and
further refined into new meaningful part denominations for the category.\\

\point{Material assignments.}
In Figure \ref{fig:materials_render}, we illustrate material categories in \compat with samples from our collection.
We collect Physics-Based Rendering (PBR~\cite{pharr_pbr_2016}) materials from various free-to-use repositories,
including the NVIDIA vMaterials\footnote{vMaterials library: \url{https://developer.nvidia.com/vmaterials}} library and the ambientCG\footnote{ambientCG public domain repository: \url{https://ambientcg.com/}} public domain material library.
We filter collected PBR materials to ensure \textbf{1)} overall visual quality,
\textbf{2)} compatibility with our rendering pipeline,
\textbf{3)} visual affinity with our collected shapes.
We collect a total of \nmaterialclassescoarse coarse material categories, for \nmaterialclasses total PBR materials. With each segmented part, a set of compatible material categories is provided by the annotators (e.g. "metal, wood" for a leg in a chair.).
The list of compatible materials for each part of each shape is first broadly defined at the 
shape category level and refined on a case-by-case basis for specific shapes.\\

\begin{figure*}
  \centering
  \vspace{-1em}
  \includegraphics[width=0.98\linewidth]{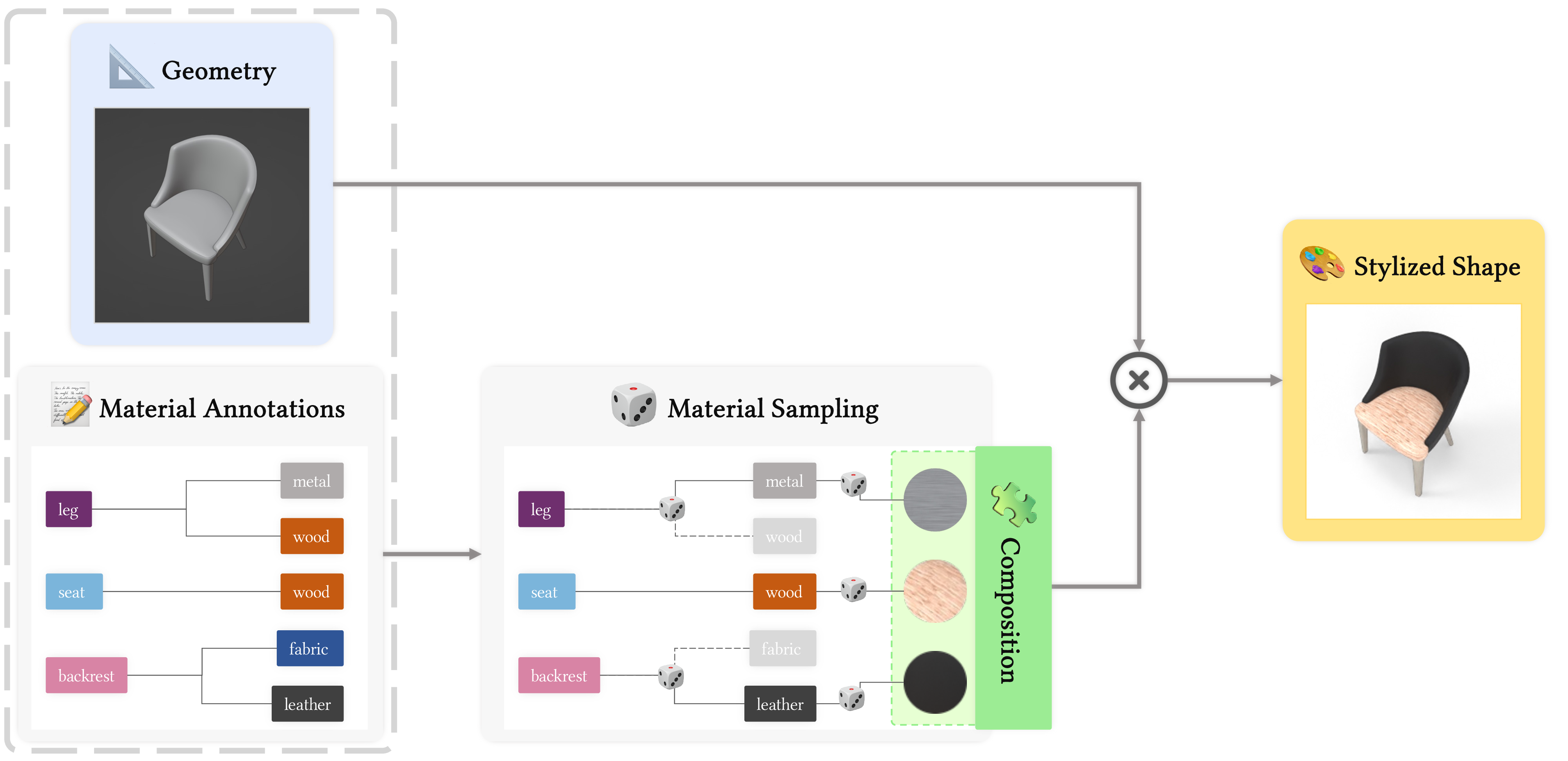}
  \caption{
    \textbf{Sampling a shape style.}
    Starting from part-material annotations, we first sample a coarse material class from a set of compatible material classes for each part.
    We then sample fine-grained materials within each sampled coarse material class. The combination of fine-grained materials for each part defines a \textit{composition}.
    Finally, we apply the sampled part-materials pairs to the shape to obtain a stylized shape.
  }
  \label{fig:style_sampling}
  \vspace{-0.5em}
\end{figure*}

\point{Stylized shapes.}
Using the collected material compatibility information associated with each part, we randomly sample a material for each part of a shape to create a \textit{style}.
A \textit{composition} is a combination of materials that could be applied to any shape, and a \textit{style} is an instance of a \textit{composition} applied to a specific shape. 
We detail the process of shape stylization in Figure \ref{fig:style_sampling}.
An average of \nstylespershape styles are sampled per shape.
The number of possible styles per shape $S$ can be defined as:

\[
  \mathcal{N}\left(S\right) \ = \prod_{p \in \mathcal{P}(S)}^{} |\mathcal{M}\left(S, p\right)|
\]

\noindent where $\mathcal{P}(S)$ denotes the set of parts belonging to shape $S$,
and $\mathcal{M}\left(S, p\right)$ the set of materials compatible with part $p$ in shape $S$.
For 14.6\% of shapes, $\mathcal{N}\left(S\right) < $ \nstylespershape,
due to either a small number of parts or compatible materials per part.
To compensate for this effect, we oversample from shapes where
$\mathcal{N}\left(S\right) >> $ \nstylespershape to reach the desired average
of \nstylespershape styles per model.

\begin{figure*}
  \centering
  \vspace{-2em}
  \includegraphics[width=0.99\linewidth]{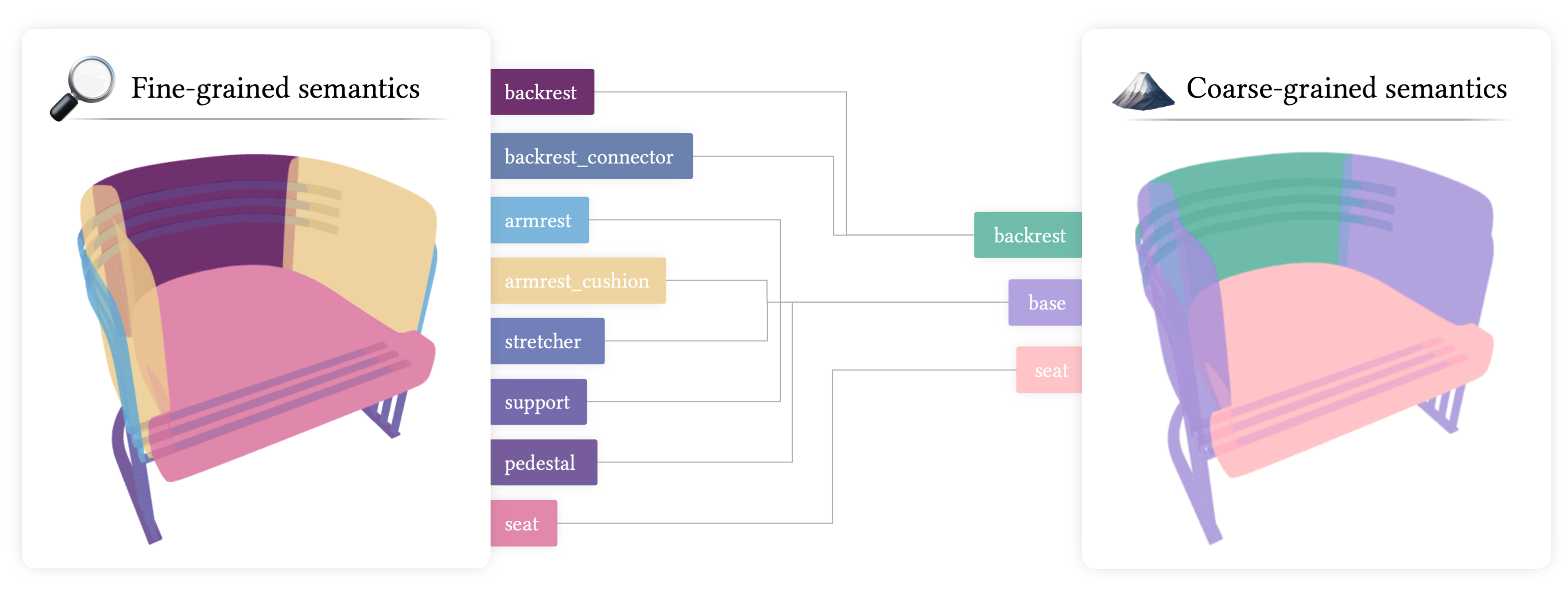}
  \vspace{0em}
  \caption{\textbf{Going from fine-grained to coarse-grained segmentation.} Fine-grained classes are merged following a shape category-specific nomenclature to create coarse-grained classes. Resulting shapes are simplified and contain fewer parts.
  }
  \label{fig:multi_levels}
  \vspace{-0.5em}
\end{figure*}

\subsection{Coarse/Fine-grained semantics} 
\noindent \compat provides part and material annotations in two hierarchical semantic levels: \textit{coarse} and \textit{fine}.\\

\point{Part hierarchies.} Fine-grained part classes are defined from a hand-defined shape category-specific nomenclature.
Coarse-grained part semantics are defined as shape category-specific groupings of fine-grained part categories.
For example, in the "\texttt{airplane}" shape category, the \texttt{wheel}, \texttt{wheel\_connector} and \texttt{wheel\_cover} \textit{fine-grained} parts are all merged into the \texttt{wheel} \textit{coarse} part.
Shape categories in our dataset can share part names by default.
Parts that are category specific and relevant to the category are prefixed by the name of the category (for example: \texttt{airplane\_wing}).

\noindent We visualize these two semantic levels in Figure \ref{fig:multi_levels} for a single 3D shape, and highlight resulting grouped parts.
The coarse-grained semantic level considerably simplifies the compositional structure of shapes, while the fine-grained semantic level provides a more detailed description of the composition of shapes.
In the coarse-grained setting, the number of shape category-specific parts is also significantly reduced, while the number of parts shared across shape categories is increased.
In Figure \ref{fig:multi_levels__page}, we provide additional examples of coarse-grained and fine-grained part semantics groupings for three distinct shape categories.
Our coarse-level part semantics can be used for tasks that require a high-level understanding of shapes, while fine-grained part semantics can be used for tasks that require more detailed, shape category-specific understanding.

\noindent We compare the average number of unique parts per object (\textbf{top}) for fine and coarse semantic levels across all shape categories in Figure \ref{fig:part_occ}.
We also plot the sorted number of occurrences of each part (\textbf{bottom}) in both semantic levels.
In the coarse-grained semantic level, parts occurrences are concentrated on a smaller number of parts, while the distribution for the fine-grained level is clearly long-tailed.
The average number of parts per object is also equalized across shape categories in the coarse-grained level, while some shape categories present a significantly higher number of parts per object in the fine-grained level like cars and bicycles.
Overall, the coarse-grained semantic level provides a more balanced distribution of parts across shape categories, while the fine-grained semantic level provides a more detailed description of the composition of shapes.\\

\point{Material hierarchies.} Coarse-grained materials correspond to a high-level set of \nmaterialclassescoarse material categories (e.g. "\texttt{wood}, \texttt{metal}, \texttt{ceramic}, etc.).
Each high-level material category is composed of fine-grained specific materials belonging to that category (e.g. "\texttt{pine\_wood}" in "\texttt{wood}".). In Table \ref{tab:material_count}, we detail the number of fine-grained materials within each coarse category in \compat.

\newcolumntype{P}[1]{>{\centering\arraybackslash}p{#1}}
\begin{table}[h]
  \caption{
      \vspace{0.5em}
      \textbf{Materials in the \compat dataset.} We show the number of fine-grained materials within each of the \nmaterialclassescoarse coarse-grained category.
  }
  \vspace{0.5em}
  \centering
  \resizebox{0.97\linewidth}{!}{
    \setlength{\extrarowheight}{0.3em}
    \begin{tabular}{@{}cP{2.8em}P{2em}P{2em}P{2em}P{2em}P{2em}P{3em}@{}}
      \toprule
      {\color{DarkOrange}{ \faPalette }} \textbf{Material} & ceramic & fabric & glass & granite & leather & marble & metal \\ \midrule
      $\mathbf{\#}$ \textbf{Count} & 6       & 32     & 6     & 11      & 34      & 39     & 66    \\ \bottomrule
      \vspace{-1em}
    \end{tabular}
  }
  \resizebox{0.97\linewidth}{!}{
    \begin{tabular}{@{}cP{2.8em}P{2em}P{2em}P{2em}P{2em}P{2em}P{3em}@{}}
      \toprule
      {\color{DarkOrange}{ \faPalette }} \textbf{Material} & plant & plastic & rubber & soil & wax & wood & \textbf{Total}  \\ \midrule
      $\mathbf{\#}$ \textbf{Count} & 1     & 21      & 5      & 4    & 4   & 64   & \textbf{\nmaterialclasses}  \\ \bottomrule
    \end{tabular}
  }
  \vspace{1em}
  \label{tab:material_count}
  \vspace{-2em}
\end{table}

\begin{figure}[h]
  \centering
  \includegraphics[width=\linewidth]{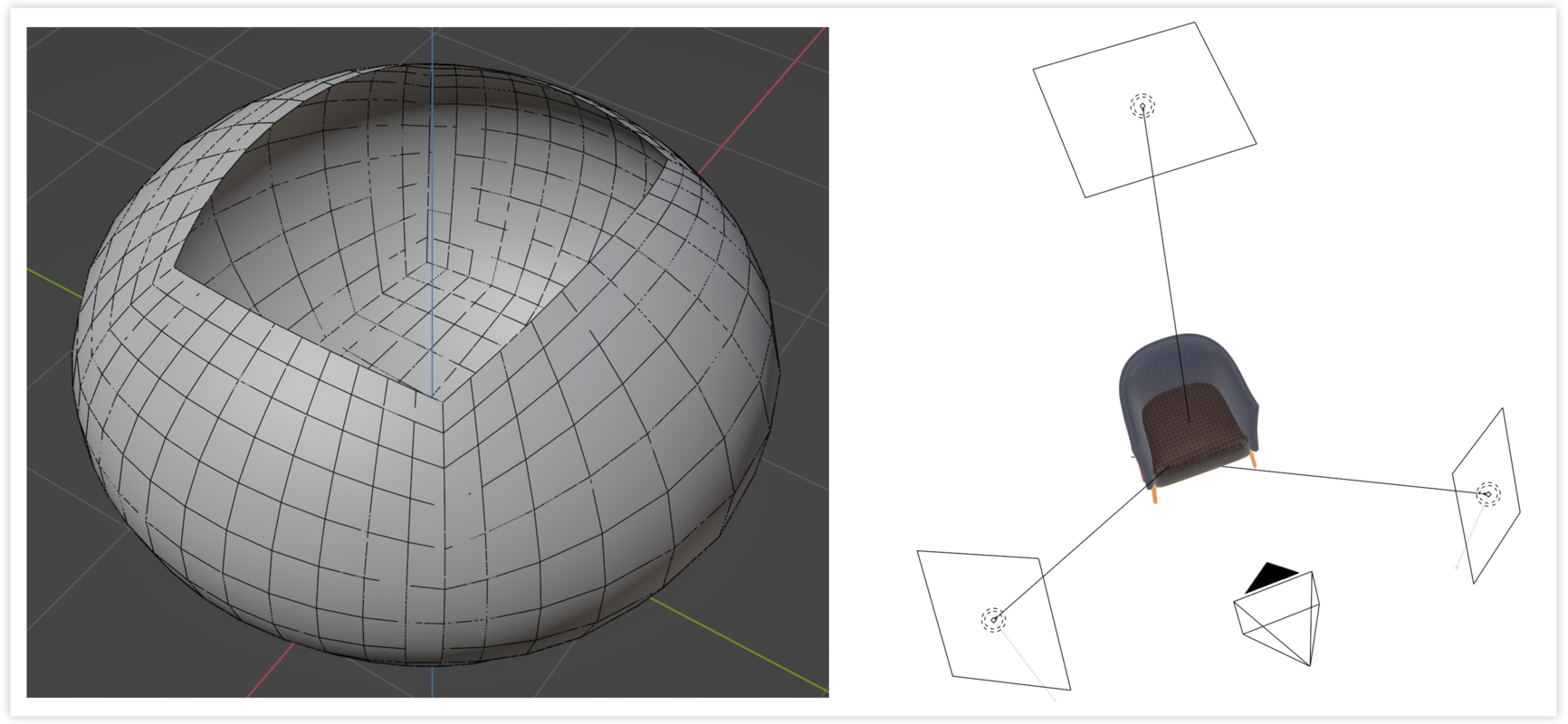}
  \vspace{-1em}
  \caption{
    \textbf{Blender rendering scene setup.}
    We depict the ovoid surface in the middle of which shapes are placed (\textbf{left}).
    We also show the scene setup used with three area lights and the camera (\textbf{right}). The directional light is positioned above the whole scene and centered on the shape.
  }
  \label{fig:render_setup}
  \vspace{-1em}
\end{figure}

\subsection{Rendering}\label{subsec:rendering}
\vspace{0.5em}
\point{Scene.} We render each shape in the same scene with a single directional light and three area lights positioned around the shape. In Figure \ref{fig:render_setup}, we detail our rendering scene setup with an example shape.
The stylized shape is placed inside an ovoid surface with a white color, to ensure that the shape is always rendered on a uniformly white background.
Projected shadows only appear on the $z=0$ plane on which the shape is placed.
When rendering depth maps and masks, the background surface is removed from the scene.

\noindent All images are rendered in a 256x256 resolution, and 2D images are encoded in the PNG format.

\noindent Depth maps are stored in the OpenEXR format to accommodate absolute distances to the image plane, which are represented using floating-point values.\\

\point{Viewpoints.} Each stylized shape is rendered from multiple perspectives: \nviewpertype canonical viewpoints and \nviewpertype random viewpoints.
We first translate each shape above the $z = 0$ plane.
Camera viewpoints are defined in spherical coordinates $(\phi, \theta)$ where the origin is set to the center of the shape's bounding box, which we note $\mathbf{o}_c$.
The camera is rotated around $\mathbf{o}_c$ by $\phi$ and $\theta$.
Canonical viewpoints are distributed evenly around the shape with a fixed elevation $\theta$.
We set the base spherical angle $\phi$ to 40 degrees and then increment it by 90 degrees for each of the four viewpoints, while keeping $\theta$ fixed at 0 degrees.
Random viewpoints are sampled uniformly from an upper hemisphere above the plane.
We randomly sample $\phi$ from the range $[0, 2\pi]$ and $\theta$ from the range $[-\frac{1}{3}\pi, \frac{1}{3}\pi]$.
Using the obtained $\phi$ and $\theta$ angles, we define the position and orientation of the camera.
The camera's initial position denoted as $\mathbf{c_0}$ is rotated around $\mathbf{o_c}$.
The orientation of the camera is then adjusted to ensure that the image plane is centered on $\mathbf{o_c}$.
Extrinsic and intrisic camera parameters are recorded for each view and are provided alongside the rendered images.
The sampling procedure of camera parameters is detailed in Algorithm \ref{alg:cam_placement}.

\subsection{Toolbox}

To support the use of \compat, we provide a toolbox for easily loading and visualizing the data.
Mainly, we provide the following elements:
\begin{itemize}
  \setlength\itemsep{0.4em}
  \item \textbf{Python API} for easily loading the data, based on PyTorch~\cite{paszke_pytorch_2019} and WebDataset~\cite{webdataset_2023}.
  \item \textbf{Web-based browser} for easily exploring 3D shapes and part annotations in both coarse and fine-grained semantic levels (see Figure \ref{fig:web_browser}).
  \item \textbf{Documentation and notebooks} to facilitate the use of the dataset.
\end{itemize}
All of these elements are available on the \compat website\footnote{\compat website: \url{https://3dcompat-dataset.org/doc}}.\\

\resizebox{0.96\linewidth}{!}{
  \hspace{-2em}
  \vspace{-2.5em}
  \begin{minipage}{\linewidth}
    \begin{algorithm}[H]
      \caption{\ Sampling camera parameters}
      \label{alg:cam_placement}
      \textbf{Input}:
      \begin{itemize}
        \item $\mathbf{obj} \in \mathbb{R}^{N \times 3}$: Object to render.
        \item $\mathbf{c_0} \in \mathbb{R}^3$: Base position of the camera.
        \item $v_k \in \llbracket 0, 3 \rrbracket $: View identifier.
        \item $\texttt{is\_random\_view} \in \{0, 1\}$: Sample a random view.
      \end{itemize}
      \textbf{Output}:
      \begin{itemize}
          \item $\mathbf{m_w} \in \mathbb{R}^{4 \times 4}$: Camera transformation matrix.
      \end{itemize}
      \begin{algorithmic}[1]
          \Function{Sample\_Camera}{$\mathbf{obj}, \mathbf{c_0}, v_k, \texttt{is\_random}$}
              \State $\mathbf{o_c} \gets \text{bounding\_box\_center}(\mathbf{obj})$
              \If{$\texttt{is\_random\_view}$}
                  \State $\mathbf{\theta} \sim \mathcal{U}{\left[ -\frac{1}{3}\pi, \frac{1}{3}\pi \right]},\ \mathbf{\phi} \sim \mathcal{U}{\left[ 0, 2\pi \right]}$
              \Else
                  \State $\mathbf{\theta} \gets 0,\ \mathbf{\phi} \gets \frac{2}{9} \pi + \frac{\pi}{2} v_k$
              \EndIf
              \State $\mathbf{c_1} \gets \text{rotate\_point}(\mathbf{c_0};\ \mathbf{o_c}, \mathbf{\phi}, \mathbf{\theta})$
              \State $\mathbf{m_w} \gets \text{look\_at}(\mathbf{c_1}, \mathbf{o_c})$
              \State \Return $\mathbf{m_w}$ \Comment{Camera transformation matrix}
          \EndFunction
      \end{algorithmic}
    \end{algorithm}
    \vspace{-0.5em}
  \end{minipage}
}

\vspace{1em}
\section{Experiments}
\subsection{Classification and Segmentation}
\vspace{0.5em}
\fpoint{Shape classification.} As illustrated in Figure \ref{fig:class_occ}, the shape class distribution of our dataset is significantly long-tailed.
We conduct shape classification experiments on 2D renders and 3D XYZ pointclouds to assess the difficulty of this task on our dataset.
All pointclouds are sampled with a resolution of 2048 points, and all methods are trained from scratch for 200 epochs.
For 2D classification, we fine-tune ResNet models~\cite{he_resnet_2015} pretrained on ImageNet~\cite{deng_imagenet_2009} for 30 epochs.
We report 2D and 3D shape classification results in Table \ref{tab:shape_classification}.
We reach a maximum top-1 accuracy of 90.20\% on 2D renders with ResNet-50, and 85.14\% on 3D pointclouds with CurveNet~\cite{xiang_curvenet_2021}.\\

\point{Part segmentation.} We conduct 3D part segmentation experiments on pointclouds and 2D renders to assess the difficulty of this task on our dataset.
We provide results for both fine-grained and coarse-grained 3D part segmentation in Table \ref{tab:3D_part_segmentation}.
We report pointwise accuracy (\textit{shape-agnostic}) and mIOU for each model.
For mIOU, we consider the shape-informed version of the metric where we restrict the set of predicted parts to the parts that are present in the ground-truth shape category, and the shape-agnostic version where all possible parts are considered.
We also report results with and without using a shape prior during training and inference for PCT~\cite{guo_pct_2021}, PointNet++~\cite{qi_pointnet_2017} and CurveNet~\cite{xiang_curvenet_2021}.
We note that getting accurate part segmentations without RGB information is challenging but remains possible.
Without using a shape prior, CurveNet~\cite{xiang_curvenet_2021} reaches a shape-agnostic mIOU of 53.09\% on fine-grained part segmentation.\\

\noindent In this setting, the model has to perform the challenging task of point-wise part classification from a set of \npartsfine possible parts.

\noindent Overall, a large gap exists (around 30 accuracy points across models) between shape-informed and shape-agnostic mIOU, highlighting the difficulty of the task over the full space of possible parts.
The task of coarse-grained part segmentation is easier, as the model only has to perform part classification from a set of \npartscoarse possible parts. In this setting, CurveNet~\cite{xiang_curvenet_2021} reaches a shape-agnostic mIOU of 76.32\%. 
For 2D fine and coarse part segmentation, we report results for SegFormer~\cite{xie_segformer_2021} in Table \ref{tab:2D_part_segmentation}, alongside material segmentation results.
We obtain a mIOU of 52.24\% for fine-grained part segmentation, 73.35\% for coarse-grained part segmentation, and 82.45\% for material segmentation.

\subsection{Grounded Compositional Recognition (GCR)}

\fpoint{Task.} One key property of our \compat dataset is that it enables understanding of the complete part-material compositions of a given 3D shape.
This involves predicting the category of the object, all part categories, and the associated materials for each of those parts within the 3D model.
In Figure \ref{fig:gcr_task}, we detail all information that has to be predicted for a given shape in this proposed GCR task.
Grounded Compositional Recognition can be related to Zero-Shot Recognition, which aims at predicting the category of an object from a set of unseen categories, where the unseen categories are defined by unseen compositions of visual attributes~\cite{lampert_learning_2009,farhadi_describing_2009,elhoseiny_write_2013}.
The GCR task can also be related to situation recognition~\cite{pratt_grounded_2020, yatskar_situation_2016} which can be defined as the identification of role - entity pairs in a given scene~\cite{pratt_grounded_2020}.\\

%==============================================================================================================
%==============================================================================================================

\point{Metrics.} Drawing inspiration from the metrics introduced in~\cite{pratt_grounded_2020, yatskar_situation_2016} initially designed for the compositional recognition of activities in images, we define the GCR compositional metrics in 2D/3D as follows:
\vspace{0.5em}
\begin{itemize}
  \setlength\itemsep{0.5em}
  \item \textbf{Shape accuracy.} Proportion of correctly predicted shape categories.
  \item \textbf{Value.} Proportion of correctly predicted part-material pairs.
  \item \textbf{Value-all.} Accuracy of predicting all part-material pairs of a shape correctly.
\end{itemize}
\vspace{0.5em}
\noindent We extend these metrics to the segmentation of parts and materials in 2D/3D by defining \textit{grounded} variants of \textbf{value} and \textbf{value-all} metrics:
\vspace{0.5em}
\begin{itemize}
  \setlength\itemsep{0.5em}
  \item \textbf{Grounded-value.} Proportion of correctly predicted part-material pairs, where the part is correctly segmented.
  \item \textbf{Grounded-value-all.} Accuracy of predicting all part-material pairs of a shape correctly, where all parts are correctly segmented.
\end{itemize}
\vspace{0.5em}
\noindent We consider a part to be correctly segmented if the predicted part segmentation mask has an intersection over union (IoU) of at least 0.5 with the ground-truth part segmentation mask.\\

\noindent Formally, for a test set with $N$ shapes, let $y_i$ and $\hat{y}_i$ denote the true and predicted shape categories for shape $i$, respectively. Additionally, let $\mathcal{C}_i = \{(p_1, m_1), (p_2, m_2), \ldots, (p_{K_i}, m_{K_i})\}$ represent the set of true part-material pairs for shape $i$, and $\hat{\mathcal{C}}_i$ the corresponding predicted pairs.
\noindent Let $\mathcal{P}_i$ denote the set of all parts in shape $i$.\\

\noindent Using these definitions, we define the metrics as follows.

\vspace{1em}
\noindent \textit{Shape Accuracy} measures the proportion of correctly identified shape categories:
$$\textsc{Shape Accuracy} = \frac{1}{N} \sum_{i=1}^{N} \mathds{1}(y_i = \hat{y}_i)$$

\vspace{1em}
\noindent \textit{Value} computes the average proportion of correctly predicted part-material pairs per shape:
$$\textsc{V} = \frac{1}{N} \sum_{i=1}^{N} \frac{|\mathcal{C}_i \cap \hat{\mathcal{C}}_i|}{|\mathcal{C}_i|}$$

\vspace{1em}
\noindent \textit{Value-all} requires perfect prediction of all part-material pairs:
$$\textsc{Vall} = \frac{1}{N} \sum_{i=1}^{N} \mathds{1}(\mathcal{C}_i = \hat{\mathcal{C}}_i)$$

\vspace{1em}
\noindent With $\mathds{1}(\cdot)$ being the indicator function. Note that performing well on these metrics only requires part-material labels to be predicted, and does not require part segmentation masks.
For the grounded variants, we define $\mathcal{S}_i$ as the set of correctly segmented parts in shape $i$, that is:

$$S_i = \{p \in \mathcal{P}_i : \text{IoU}(p, \hat{p}) \geq 0.5\}$$

\noindent The $\text{IoU}(p, \hat{p})$ function computes the intersection over union between the predicted part $\hat{p}$ and the ground-truth part $p$.
Depending on the modality, the $\text{IoU}(\cdot)$ function is defined either as the pixelwise IoU for 2D images or the pointwise IoU for 3D pointclouds.
This allows us to define grounded variants of the metrics as follows:

\vspace{1em}
\noindent \textit{Grounded-value} which measures correctly predicted part-material pairs that are also correctly segmented:
$$\textsc{GV} = \frac{1}{N} \sum_{i=1}^{N} \frac{|\{(p, m) \in \mathcal{C}_i : p \in \mathcal{S}_i \land (p, m) \in \hat{\mathcal{C}}_i\}|}{|\mathcal{C}_i|}$$

\vspace{1em}
\noindent \textit{Grounded-value-all} which requires all part-material pairs to be correctly predicted and all parts to be correctly segmented:
$$\textsc{GVall} = \frac{1}{N} \sum_{i=1}^{N} \mathds{1}\left((\mathcal{C}_i = \hat{\mathcal{C}}_i) \land (\forall p \in \mathcal{P}_i: p \in S_i) \right)$$

\noindent Note that \textit{Value} and \textit{Grounded-value} are both evaluated at the shape level: we divide the number of correctly identified (resp. grounded) part-material pairs by the total number of parts appearing in each shape, and then average across all samples.
\textit{Value-all} is thus upper bounded by \textit{Value}, and \textit{Grounded-value-all} by \textit{Grounded-value}.\\

%==============================================================================================================
%==============================================================================================================

% ======================================================
% ======================================================
\cleardoublepage
\clearpage

\begin{table*}
  \vspace{-2.5em}
  \centering
  \caption{
    \vspace{0.5em}
    \textbf{3D Part segmentation.}
    We report mean intersection over union (mIOU) and pointwise accuracy for various models
    on the \compat dataset for both fine-grained (\textbf{left}) and coarse-grained (\textbf{right}) 3D part segmentation.
    For mIOU, we differentiate between shape-informed (where the ground-truth shape category is provided as input) and shape-agnostic evaluation.
    For PCT, PointNet++ and CurveNet, we also report results with and without using a shape prior during training and inference.
    All models are trained on 3D XYZ pointclouds.
  }
  \vspace{0.5em}
  \setlength{\extrarowheight}{0.3em}
  \resizebox{0.495\linewidth}{!}{
    \begin{tabular}{@{}lcccc@{}}
      \toprule
      \multicolumn{1}{c}{\multirow{2}{*}{\textbf{Model}}} & \multicolumn{1}{c}{\multirow{2}{*}{\textbf{Shape Prior}}} & \multirow{2}{*}{\textbf{Pointwise Acc. (\%)}} & \multicolumn{2}{c}{\textbf{mIOU (\%)}}                 \\ \cmidrule(l){4-5} 
      \multicolumn{1}{c}{}                                & \multicolumn{1}{c}{}                                      &                                          & \textbf{Shape-Informed} & \textbf{Shape-Agnostic} \\ \midrule
      \multirow{2}{*}{PCT~\cite{guo_pct_2021}}              & \icono  & 70.49 & 81.31 & 49.09 \\
                                                            & \icoyes & 78.51 & 82.84 & 56.14 \\ \cmidrule(l){2-5} 
      \multirow{2}{*}{PointNet++~\cite{qi_pointnet_2017}}   & \icono  & 71.09 & 80.01 & 50.39 \\
                                                            & \icoyes & 78.61 & 81.19 & 56.46 \\ \cmidrule(l){2-5} 
      \multirow{2}{*}{CurveNet~\cite{xiang_curvenet_2021}}  & \icono  & 72.49 & 81.37 & 53.09 \\
                                                            & \icoyes & 79.90 & 82.15 & 59.61 \\ \cmidrule(l){2-5} 
      PointNeXt~\cite{qian_pointnext_2022}                  & \icoyes & 82.07 & 83.92 & 63.73 \\ 
      PointStack~\cite{wijaya_pointstack_2022}              & \icoyes & 78.51 & 81.98 & 56.20 \\ \midrule
      \multicolumn{5}{c}{\multirow{2}{*}{\textbf{Fine-grained segmentation}}} \\
      \multicolumn{5}{c}{}
    \end{tabular}
  }\hspace{0.25em}
  \resizebox{0.495\linewidth}{!}{
    \begin{tabular}{@{}lcccc@{}}
      \toprule
      \multicolumn{1}{c}{\multirow{2}{*}{\textbf{Model}}} & \multicolumn{1}{c}{\multirow{2}{*}{\textbf{Shape Prior}}} & \multirow{2}{*}{\textbf{Pointwise Acc. (\%)}} & \multicolumn{2}{c}{\textbf{mIOU (\%)}}                 \\ \cmidrule(l){4-5} 
      \multicolumn{1}{c}{}                                & \multicolumn{1}{c}{}                                      &                                          & \textbf{Shape-Informed} & \textbf{Shape-Agnostic} \\ \midrule
      \multirow{2}{*}{PCT~\cite{guo_pct_2021}}             & \icono  & 80.64 & 75.49 & 66.95 \\
                                                           & \icoyes & 92.15 & 82.57 & 80.49 \\ \cmidrule(l){2-5} 
      \multirow{2}{*}{PointNet++~\cite{qi_pointnet_2017}}  & \icono  & 84.73 & 77.98 & 73.79 \\
                                                           & \icoyes & 92.02 & 81.82 & 81.03 \\ \cmidrule(l){2-5} 
      \multirow{2}{*}{CurveNet~\cite{xiang_curvenet_2021}} & \icono  & 86.02 & 80.64 & 76.32 \\
                                                           & \icoyes & 93.40 & 84.62 & 83.85 \\ \cmidrule(l){2-5} 
      PointNeXt~\cite{qian_pointnext_2022}                 & \icoyes & 94.18 & 86.80 & 85.46 \\
      PointStack~\cite{wijaya_pointstack_2022}             & \icoyes & 93.49 & 84.97 & 83.67 \\ \midrule
      \multicolumn{5}{c}{\multirow{2}{*}{\textbf{Coarse-grained segmentation}}} \\
      \multicolumn{5}{c}{}
    \end{tabular}
  }
  \label{tab:3D_part_segmentation}
  \vspace{-2em}
\end{table*}

\begin{table*}
  \vspace{2em}
  \centering
  \begin{minipage}[t]{0.48\textwidth}
    \centering
    \setlength{\extrarowheight}{0.3em}
    \caption{
      \vspace{0.5em}
      \textbf{2D Part segmentation.}
      We report mean intersection over union (mIOU) for SegFormer on three 2D part segmentation tasks: fine-grained part segmentation, coarse-grained part segmentation and material segmentation.
    }
    \vspace{0em}
    \resizebox{\linewidth}{!}{
      \begin{tabular}{@{}ccc@{}}
        \toprule
        \textbf{Model}             & \textbf{Task} & \textbf{mIOU (\%)} \\ \midrule
        \multirow{3}{*}{SegFormer~\cite{xie_segformer_2021}} & Fine-grained part segmentation      & 52.24    \\
                                                             & Coarse-grained part segmentation    & 73.35    \\
                                                             & Material segmentation               & 82.45    \\ \bottomrule
        \end{tabular}
    }
    \label{tab:2D_part_segmentation}
  \end{minipage}%
  \begin{minipage}[t]{0.48\textwidth}
    \centering
    \setlength{\extrarowheight}{0.3em}
    \caption{
      \vspace{0.5em}
      \textbf{Shape classification.}
      We report the accuracy of various models on the \compat dataset for both 3D and 2D shape classification.
      3D models are trained on pointclouds provided with each shape, while 2D models are trained and tested on renders with canonical and random viewpoints.
    }
    \vspace{0em}
    \resizebox{0.95\linewidth}{!}{
    \begin{tabular}{@{}ll@{}}
        \begin{tabular}{@{}lc@{}}
          \toprule
          \multicolumn{1}{c}{\multirow{2}{*}{\textbf{Model}}} & \textbf{2D Classification}    \\ \cmidrule(l){2-2} 
          \multicolumn{1}{c}{}                                & \textbf{Accuracy (top-1, \%)} \\ \midrule
          ResNet-18~\cite{he_resnet_2015}                     & 76.27                         \\
          ResNet-50~\cite{he_resnet_2015}                     & 90.20                         \\ \bottomrule
        \end{tabular}
        & 
        \begin{tabular}{@{}lc@{}}
          \toprule
          \multicolumn{1}{c}{\multirow{2}{*}{\textbf{Model}}} & \textbf{3D Classification}    \\ \cmidrule(l){2-2} 
          \multicolumn{1}{c}{}                                & \textbf{Accuracy (top-1, \%)} \\ \midrule
          PCT~\cite{guo_pct_2021}                             & 68.88                         \\
          DGCNN~\cite{wang_dgcnn_2019}                        & 78.85                         \\
          PointNet++~\cite{qi_pointnet_2017}                  & 84.10                         \\
          PointStack~\cite{wijaya_pointstack_2022}            & 83.04                         \\
          CurveNet~\cite{xiang_curvenet_2021}                 & 85.14                         \\
          PointNeXt~\cite{qian_pointnext_2022}                & 82.21                         \\
          PointMLP~\cite{ma_pointmlp_2022}                    & 83.71                         \\ \bottomrule
        \end{tabular}
    \end{tabular}
    }
    \label{tab:shape_classification}
  \end{minipage}
  \vspace{0em}
\end{table*}

\begin{figure*}
  \vspace{0.25em}
  \centering
  {
  %include the SVG
  \includesvg[width=\linewidth]{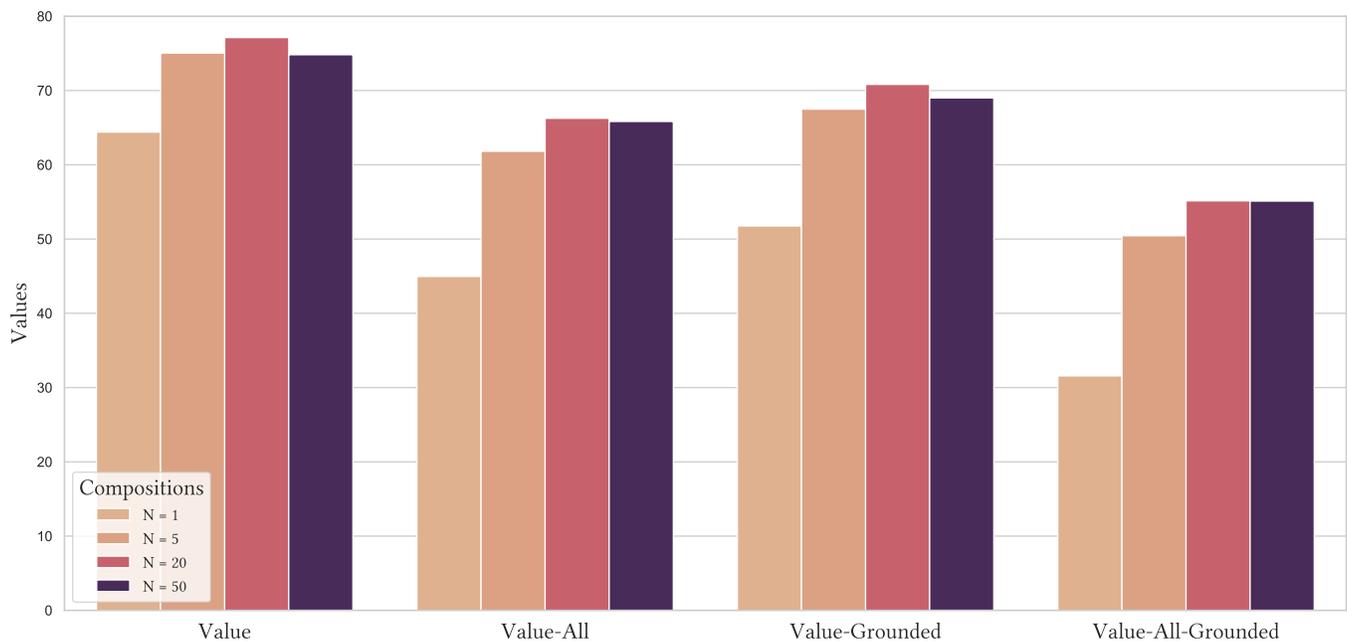}
  }
  \vspace{-2em}
  \caption{
    \textbf{Analysis of the performance with different numbers of sampled styles.}
    We train the BPNet~\cite{hu_bpnet_2021} model using 1/5/20/50 style compositions and report all the compositional metrics defined in Figure \ref{fig:gcr_task}.
    Overall, we observe a clear trend of improvement with the number of styles, especially for the \textbf{Value-All} and \textbf{Value-All-Grounded} metrics.
    We do notice however the start of a saturation effect when training with $N=50$ styles.
  }
  \label{fig:comps_ablation}
  \vspace{-0em}
\end{figure*}

\begin{table*}
    \vspace{1em}
  \centering
  \caption{
    \vspace{0.5em}
    \textbf{Grounded Compositional Recognition (GCR).}
    We evaluate the performance of various baselines on the GCR task.
    While modality fusion-based methods like BPNet~\cite{hu_bpnet_2021} and PointNeXt+SegFormer~\cite{qian_pointnext_2022,xie_segformer_2021} perform well on the shape classification task they still underperformed compared to the RGB pointcloud-based baseline.
    We report the GCR metrics under both fine-grained and coarse-grained settings, using 10 compositions per shape.
    %Overall, recognizing and grouding all part-material pairs of a shape is particularly challenging, especially in the fine-grained setting.
  }
  \vspace{0.5em}
  \setlength{\extrarowheight}{0.3em}
  \resizebox{\linewidth}{!}{
    \begin{tabular}{@{}llccccc@{}}
      \toprule
      \multicolumn{1}{c}{\textbf{Semantic level}} & \multicolumn{1}{c}{\textbf{Model}} & \textbf{Shape Acc.} & \textbf{Value} & \textbf{Value-all} & \textbf{Grounded-value} & \textbf{Grounded-value-all} \\ \midrule
      \multirow{3}{*}{\textit{Fine-grained}}      
                                                  & PointNeXt+SegFormer~\cite{qian_pointnext_2022,xie_segformer_2021} & \textbf{84.18}  & 42.57           & 9.05            & 26.68           & 3.84   \\
                                                  & BPNet~\cite{hu_bpnet_2021}                                        & 79.57           & \textbf{59.98}  & \textbf{27.74}  & 45.46           & 15.41   \\
                                                  & PointNet++\textsuperscript{\textbf{RGB}}~\cite{qi_pointnet_2017}  & 83.70           & 57.78           & 25.36           & \textbf{49.34}  & \textbf{17.55}  \\ \cmidrule(l){2-7} 
      \multirow{3}{*}{\textit{Coarse-grained}}    
                                                  & PointNeXt+SegFormer~\cite{qian_pointnext_2022,xie_segformer_2021} & 84.27           & 65.61           & 44.82           & 52.82           & 29.74  \\
                                                  & BPNet~\cite{hu_bpnet_2021}                                        & 84.72           & 75.04           & 61.81           & 67.49           & 50.44  \\
                                                  & PointNet++\textsuperscript{\textbf{RGB}}~\cite{qi_pointnet_2017}  & \textbf{85.19}  & \textbf{75.66}           & \textbf{63.88}  & \textbf{72.14}  & \textbf{58.99}  \\ \bottomrule
    \end{tabular}
  }
  \label{tab:gcr_results}
\end{table*}

\clearpage
\cleardoublepage

%==============================================================================================================
%==============================================================================================================

\vspace{0.5em}
\point{Baselines.} We experiment with two fusion-based baselines to assess the performance of the GCR task on \compat.
\vspace{0.5em}
\begin{itemize}
  \setlength\itemsep{1em}
  \item \textbf{``PointNeXt+SegFormer''}.  This baseline employs separate 2D/3D models and fuses predictions at evaluation time.
  We use PointNeXt~\cite{qian_pointnext_2022} for 3D shape classification and SegFormer~\cite{xie_segformer_2021} for 2D material segmentation and 2D part segmentation.
  2D dense predictions are projected to the 3D space using the depth maps and camera parameters.
  We use this baseline to assess the feasibility of the GCR task on \compat when all part-pair predictions are performed on the 2D space.
  \item \textbf{BPNet.} We adapt the BPNet 2D/3D multimodal method to the GCR task.
  BPNet leverages complementary information from 2D and 3D modalities by fusing features from both modalities using a bidirectional projection module for feature fusion.
  We detail the BPNet architecture we employ in Figure \ref{fig:bpnet_architecture} in the appendix.
\end{itemize}

\vspace{1.5em}
\point{Challenge.} We organized a compositional 3D visual understanding challenge on the GCR task of \compat, with the goal of benchmarking the performance of various methods, in the context of the C3DV workshop at CVPR 2023 \footnote{C3DV Workshop: \url{https://3dcompat-dataset.org/workshop/C3DV23/}}.
The best-performing method (\textbf{PointNet++\textsuperscript{\textbf{RGB}}} in Table~\ref{tab:gcr_results}) on the GCR task consisted of an unimodal 3D model based on a modified PointNet++~\cite{qi_pointnet_2017} trained on 6D inputs (XYZ coordinates and RGB color) \footnote{Winning solution repository: \url{https://github.com/Cattalyya/3DCoMPaT-challenge}}. One important design choice is the point grouping method employed which relies on spatial proximity only. The winning method achieved a \textbf{Grounded-value-all} accuracy of 58.99\% in the coarse-grained setting and 17.55\% in the fine-grained setting. Other solutions included a late fusion of 2D and 3D features by averaging logits of part and material segmentation and training a PointNet++ model with additional 2D segmentation features.
More information about the challenge submissions can be found on the workshop website.\\

\point{Results.} Table \ref{tab:gcr_results} summarizes GCR results of baseline methods and challenge winners.
The PointNeXt+SegFormer 2D-based baseline is markedly outperformed by the BPNet multimodal baseline, which can be imputed to the absence of explicit 3D-aware modality fusion during training.
The winning method \textbf{PointNet++\textsuperscript{\textbf{RGB}}}, which takes only 3D point clouds as inputs and leverages a powerful point grouping module, performs best in the coarse-grained setting, reaching 58.99\% on the \textbf{Grounded-value-all} metric.
BPNet on the other hand, performs best in the fine-grained setting, and outperforms PointNet++\textsuperscript{\textbf{RGB}} on the classification metrics (Value and Value-all), but is overtaken by PointNet++\textsuperscript{\textbf{RGB}} on the grounded metrics.
More importantly, we notice that even our best performing models struggle to perform on the fine-grained GCR task, where we reach at most 17.55\% accuracy on the \textbf{Grounded-value-all} metric.
Overall, recognizing and grouding all part-material pairs of a shape is particularly challenging, especially in the fine-grained setting.
%This suggests that training a model able to achieve strong performance across GCR metrics is challenging, especially in the fine-grained setting.\\

\vspace{-0.5em}\noindent In this sense, Grounded Compositional Recognition is a challenging task that can be used to benchmark the compositional understanding of future multimodal models.\\

\point{Number of compositions.} We conduct further ablation analysis to investigate the impact of varying the number of compositions during the training of the BPNet~\cite{hu_bpnet_2021} model. We focus our analysis on the compositional metrics outlined in Figure \ref{fig:comps_ablation}, related to the GCR task (2D/3D material mIoU, 2D shape accuracy, and 3D part mIoU). We train multiple instance of the BPNet model with 1/5/20/50 compositions from each shape and report the performance obtained for each epoch, for each GCR compositional metric. Our findings reveal a clear and consistent improvement in the performance of all metrics as the number of compositions utilized in training is increased, specifically when going from $N_c=1$ to $N_c=5$ and $N_c=20$. However, the observed trend becomes less discernible when transitioning from $N_c=20$ to $N_c=50$ compositions. This highlights the need for further investigation into efficient ways of leveraging a large number of compositions during training.\\

\section{Conclusion}
We introduce \compat, a large-scale dataset of Compositions of Materials on Parts of 3D Things, which contains \nstylizedshapesshort styled models stemming from \nshapes 3D shapes from \nshapeclasses object categories. \compat contains 3D shapes, part segmentation information in fine-grained and coarse-grained semantic levels and material compatibility information, so that multiple high-quality PBR materials can be assigned to the same shape part.
We also propose a new task, dubbed as Grounded CoMPaT Recognition (GCR), that our dataset enables and introduces baseline methods to solve them.
Future directions may include additional tasks such as 3D part-aware shape synthesis, 3D part-aware reconstruction from 2D views, and 3D part-style transfer, which all can be enabled by the rich data provided in \compat.

\section*{Acknowledgements}
For computing support, this research used the resources of the Supercomputing Laboratory at King Abdullah University of Science \& Technology (KAUST). We extend our sincere gratitude to the KAUST HPC Team for their invaluable assistance and support during the course of this research project.
We also thank the Amazon Open Data program for providing us with free storage of our large-scale data on their servers, and the Polynine team for their relentless effort in collecting and annotating the data.

\cleardoublepage
\appendices
\section{}
\begin{figure*}
  \vspace{-1em}
  \centering
  \subfloat{%
    \includegraphics[height=0.4\linewidth]{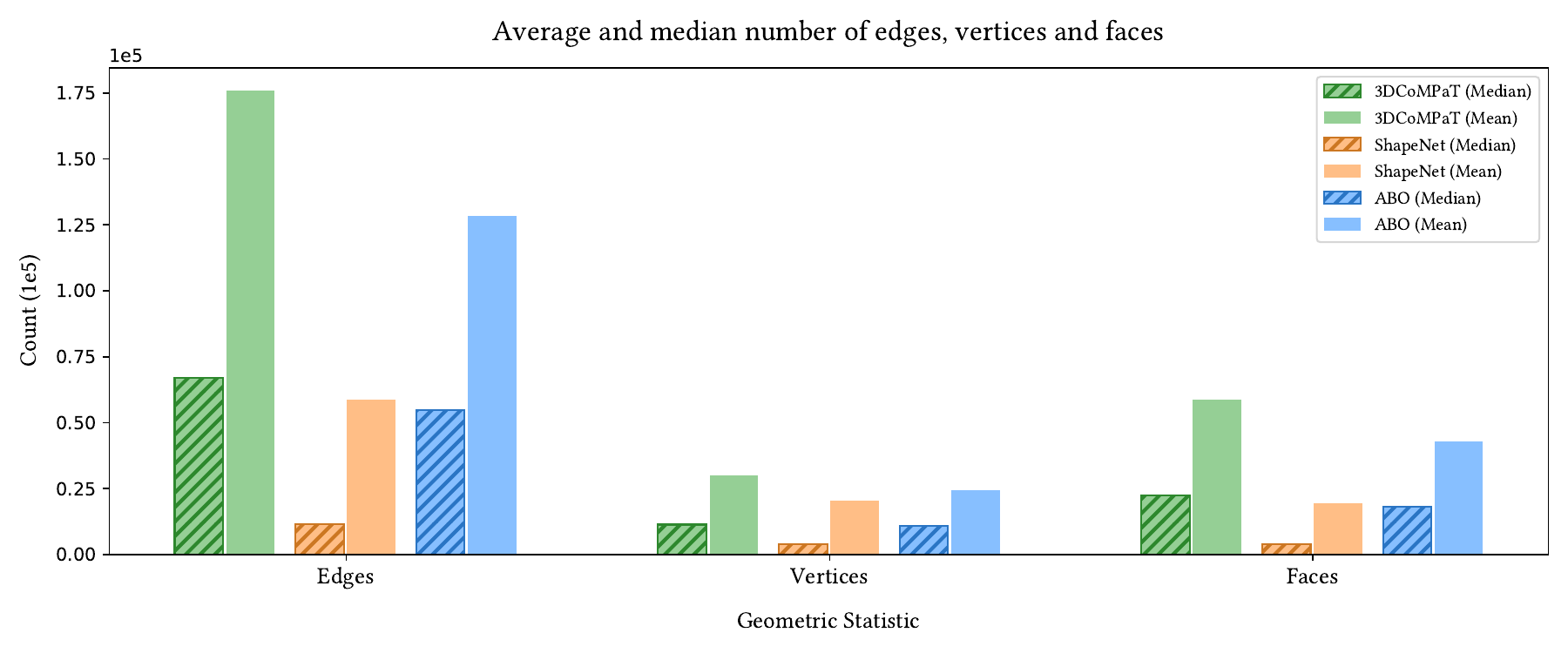}
  }
  \\
  \subfloat{%
    \includegraphics[height=0.4\linewidth]{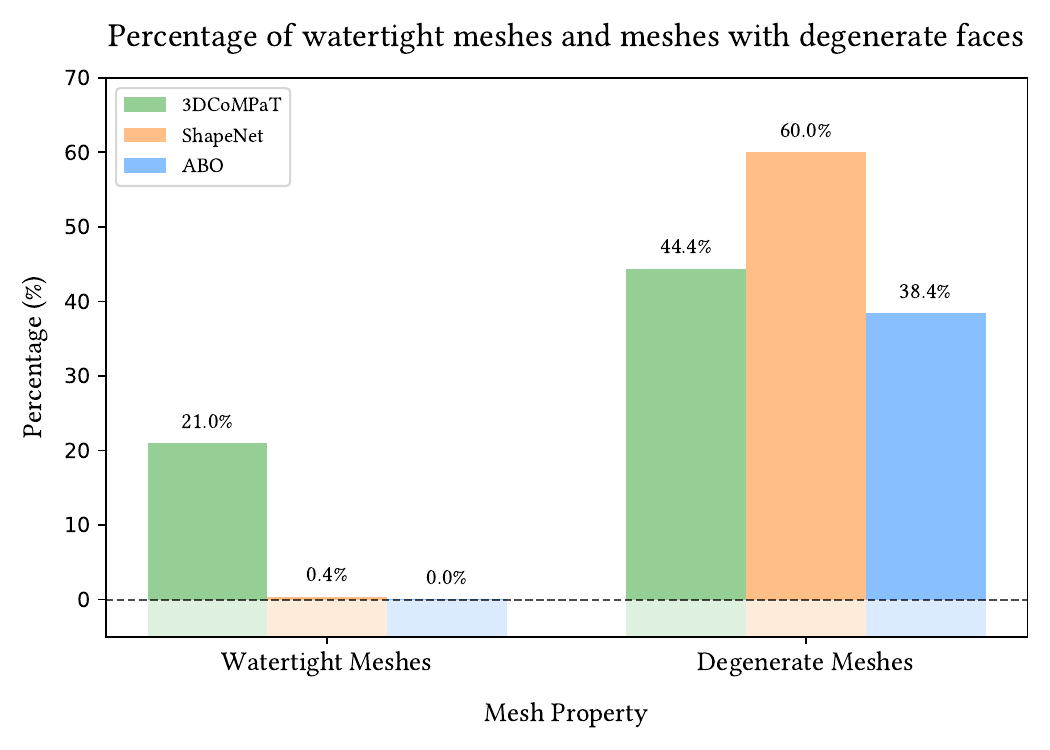}
  }
  \vspace{0em}
  \caption{\textbf{Mesh quality statistics.} Statistics on the mesh quality of \compat compared to the original ShapeNet~\cite{chang_shapenet_2015} dataset, and the ABO~\cite{collins_abo_2022} dataset.\\
  \textbf{Top:} Number of vertices, faces and edges per 3D shape.
  \textbf{Bottom:} Percentage of meshes with non-manifold geometries and percentage of meshes with degenerate meshes.
  Overall, we observe that \compat has a higher percentage of watertight meshes and lower percentage of degenerate meshes compared to the original ShapeNet dataset.
  Our dataset also compares favorably to the ABO dataset, which contains almost only meshes with non-manifold geometries.
  }
  \label{fig:mesh_quality}
\end{figure*}

Additional mesh quality statistics compared to the original ShapeNet~\cite{chang_shapenet_2015} dataset and the ABO~\cite{collins_abo_2022} dataset are provided in Figure \ref{fig:mesh_quality}, including the number of vertices, faces, and edges per shape, percentage of watertight meshes and percentage of meshes with degenerate faces.

\section{}
\begin{figure*}
  \vspace{-2em}
  \centering
  \includegraphics[width=0.98\textwidth]{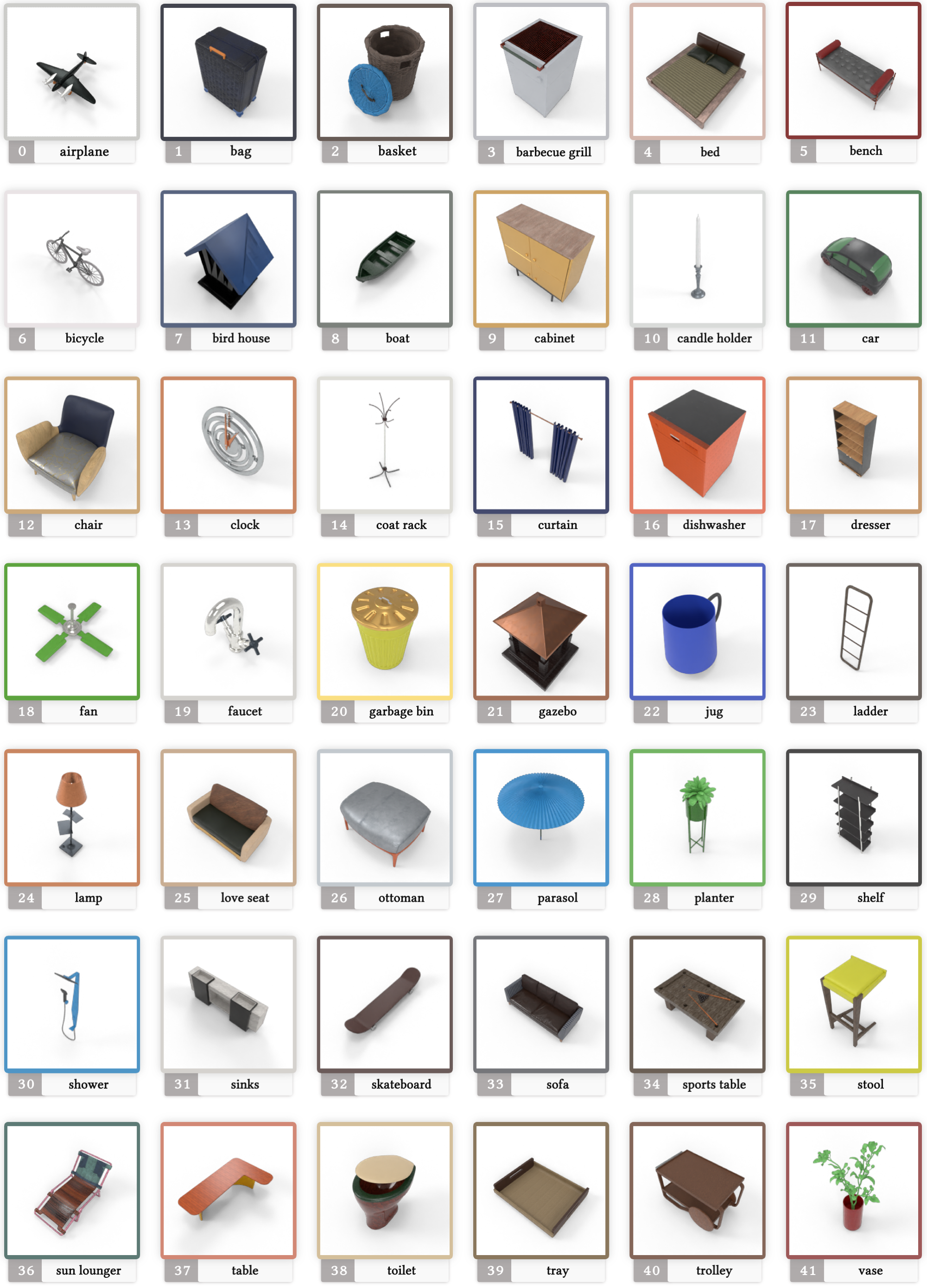}
  \vspace{0.5em}
  \caption{
  \textbf{Rendered shapes from \compat.} We render representative shapes from every \nshapeclasses shape category in \compat, with the shape category indexes and names.
  }
  \label{fig:all_classes}
  \vspace{-1em}
\end{figure*}

Rendered representative shapes from every \nshapeclasses shape category in \compat in Figure \ref{fig:all_classes}.

\section{}
\begin{figure*}
  \vspace{-2em}
  \centering
  \includegraphics[width=0.98\textwidth]{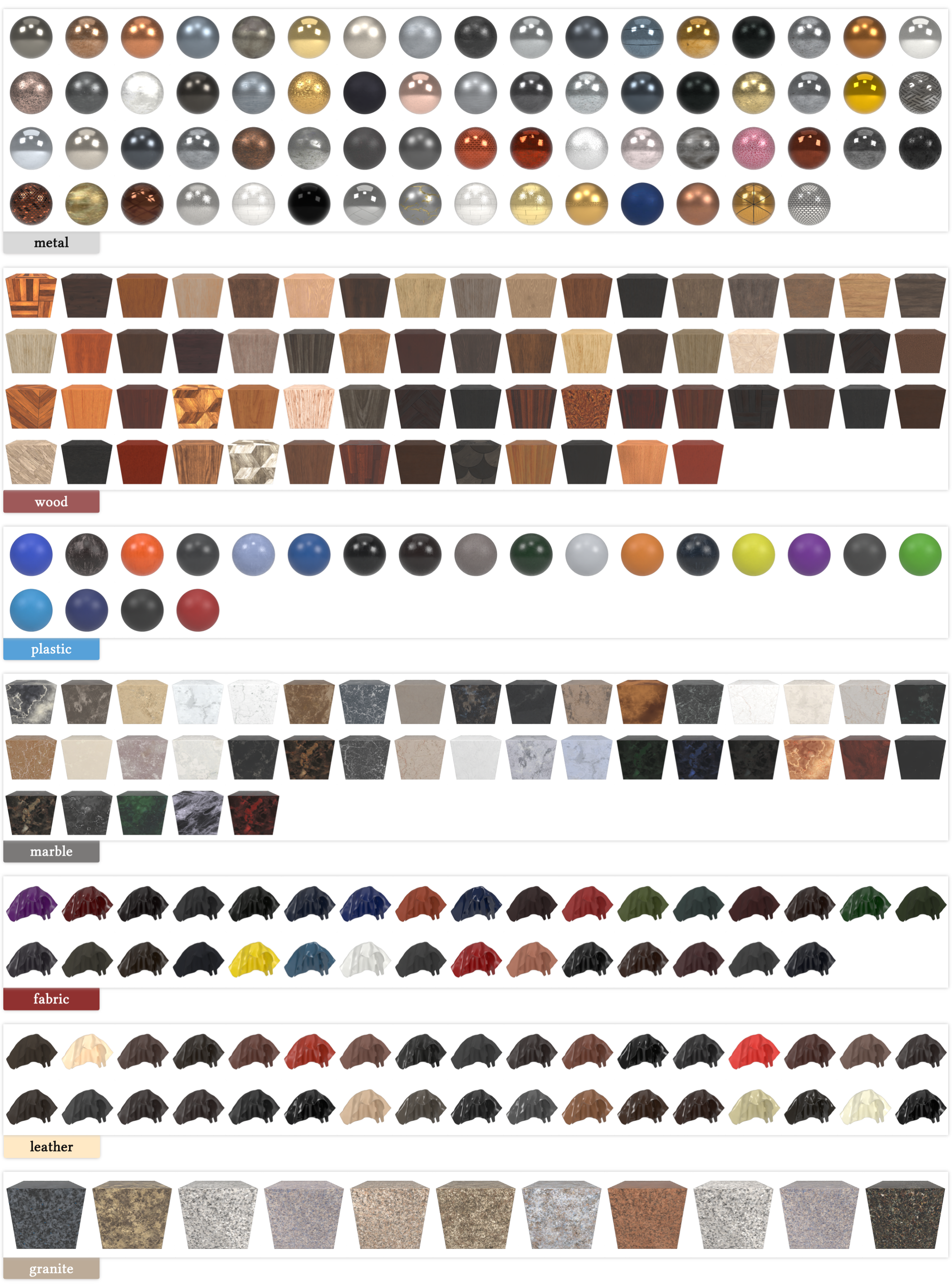}
  \vspace{0.5em}
  \caption{
  \textbf{Rendered materials.} We render the material collection we use from every part-material assignment in our dataset. We display \nmaterialclasses fine-grained material categories, grouped by coarse material category (continued in the next page).
  }
  \label{fig:all_materials}
  \vspace{-1em}
\end{figure*}

Rendered materials from every \nmaterialclasses material category in \compat in Figure \ref{fig:all_materials}.

\section{}
\begin{figure*}
  \centering
  \includegraphics[width=\linewidth]{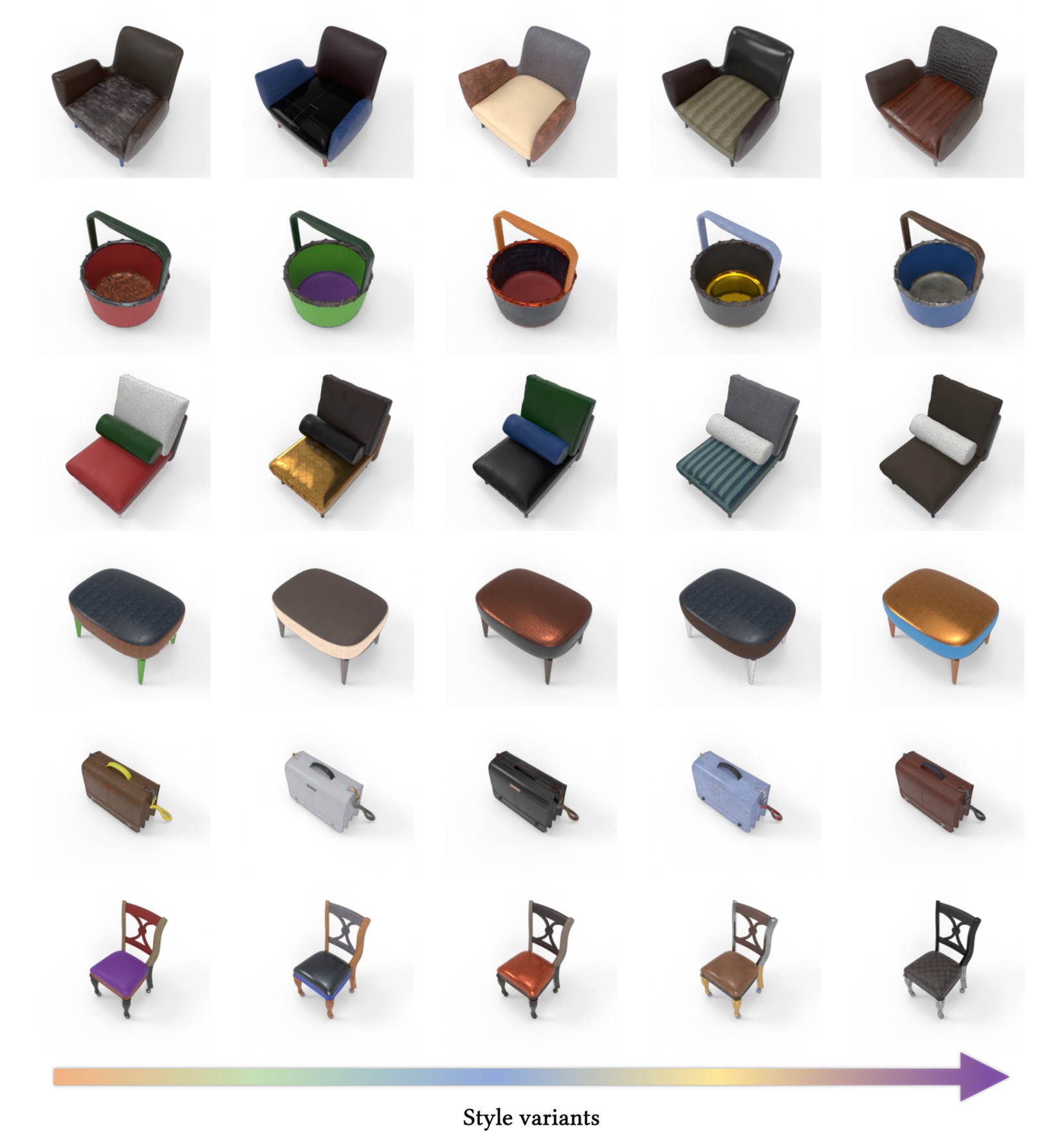}
  \caption{\textbf{Generated styles.} Randomly sampled styles for a set of six geometries from the \compat dataset.
  Styles are generated by sampling a set of compatible materials from the material collection, using the style generation algorithm described in Figure \ref{fig:style_sampling}.
  }
  \label{fig:style_variants}
\end{figure*}

Example style variants for a few randomly sampled geometries in Figure \ref{fig:style_variants}.

\section{}
\begin{figure*}
  \centering
  \vspace{-2em}
  \includegraphics[width=0.95\textwidth]{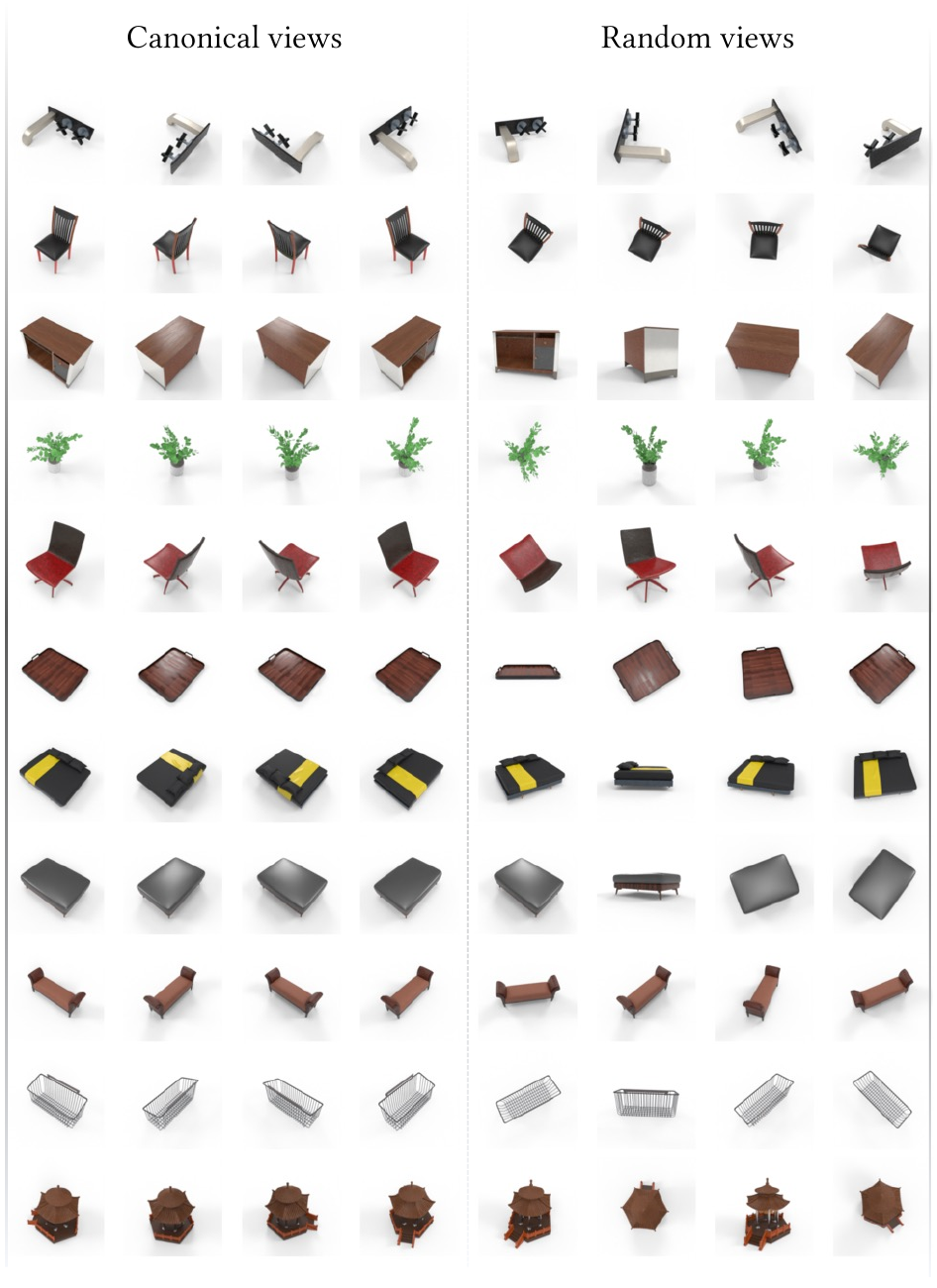}
  \vspace{0em}
  \caption{\textbf{Canonical and random views.}
  Canonical (\textbf{first four columns}) and random views (\textbf{last four columns}) rendered for various stylized shapes and shape categories.
  Random viewpoints are sampled, while canonical viewpoints are equally spaced around the shape.}
  \label{fig:views_additional}
\end{figure*}

Additional examples for the multiple views rendered per stylized shape are provided in Figure \ref{fig:views_additional}.

\section{}
\begin{figure*}
  \centering
  \includegraphics[width=\linewidth]{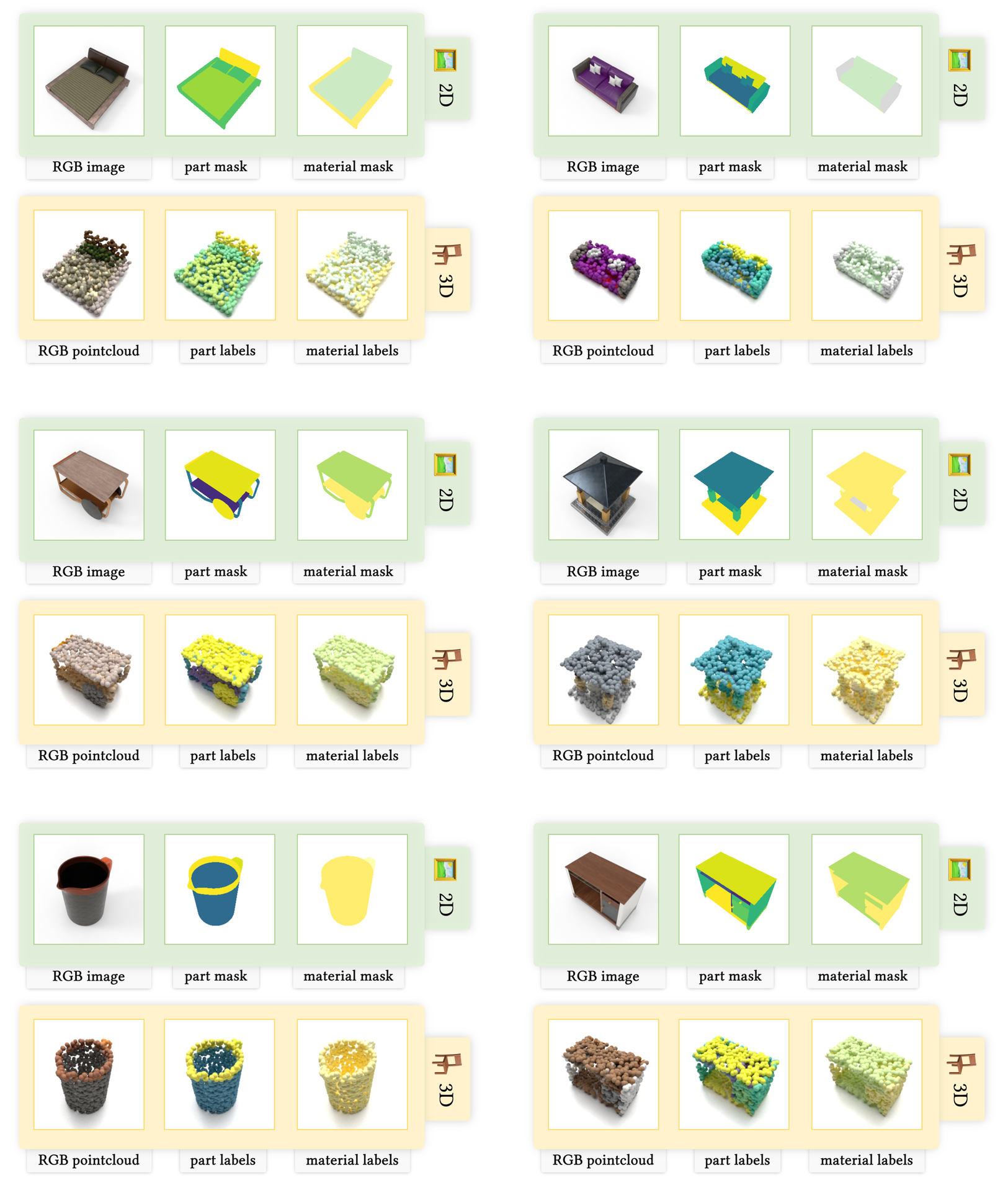}
  \caption{\textbf{Complete data samples.}
  Additional examples for data provided for six diverse shapes, following the first canonical viewpoint.
  Each 2D image is complemented by corresponding part segmentation masks, material masks, and absolute depth maps.
  Camera parameters are also provided for each render.
  For each stylized shape, we also provide RGB pointclouds labeled at the part and material levels.
  Part segmentation data is available in both \textit{coarse-grained} and \textit{fine-grained} resolutions.
  }
  \label{fig:full_data_example_appendix}
\end{figure*}

Additional examples for 2D/3D data provided with each stylized shape are provided in Figure \ref{fig:full_data_example_appendix}.

\begin{figure*}
  \vspace{-2em}
  \centering
  \includegraphics[width=0.98\textwidth]{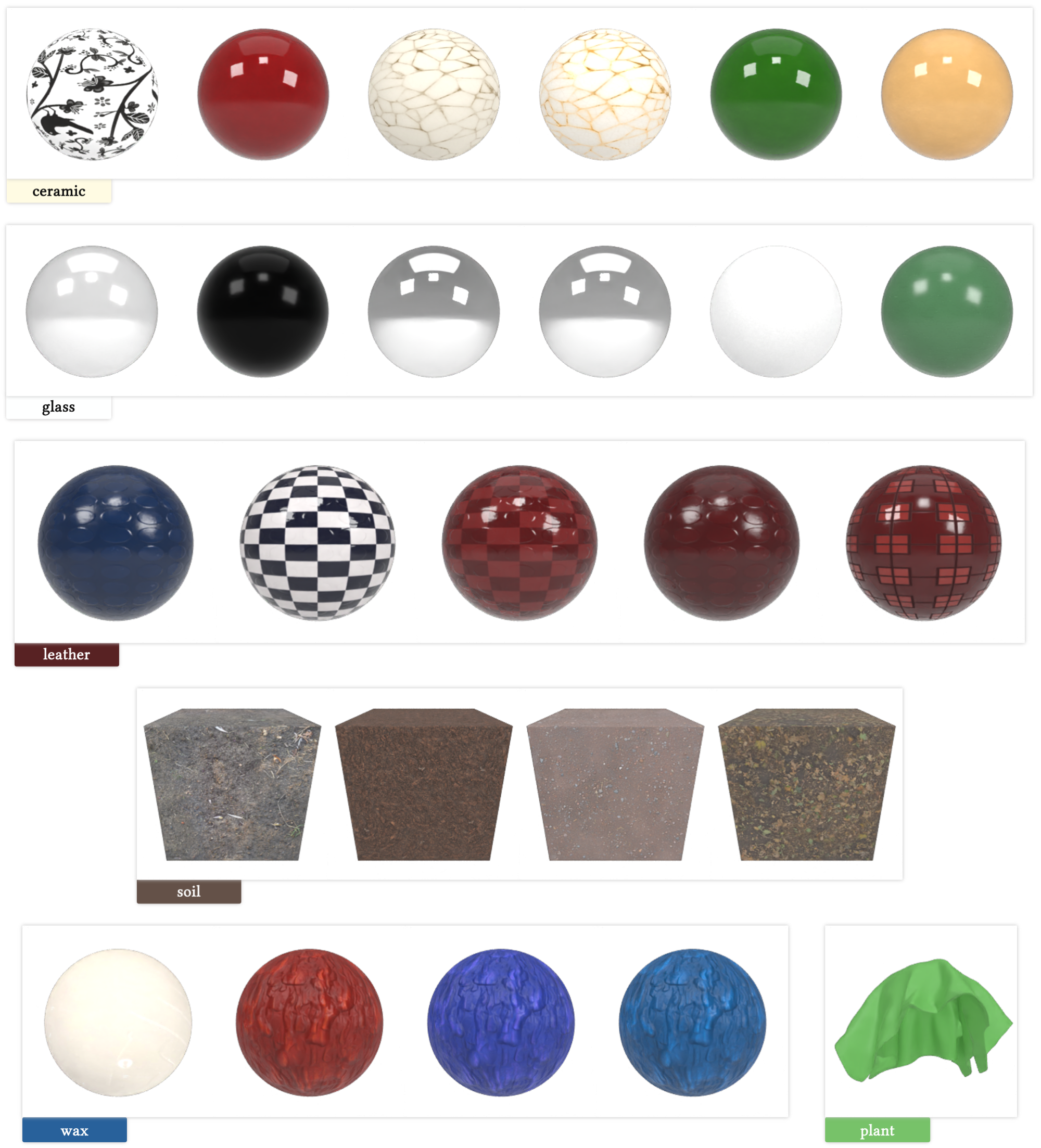}
  \vspace{0.5em}
  \label{fig:all_materials__02}
  \vspace{-1em}
\end{figure*}

\section{}
\begin{figure*}
  \centering
  \vspace{-2em}
  \includegraphics[width=0.98\linewidth]{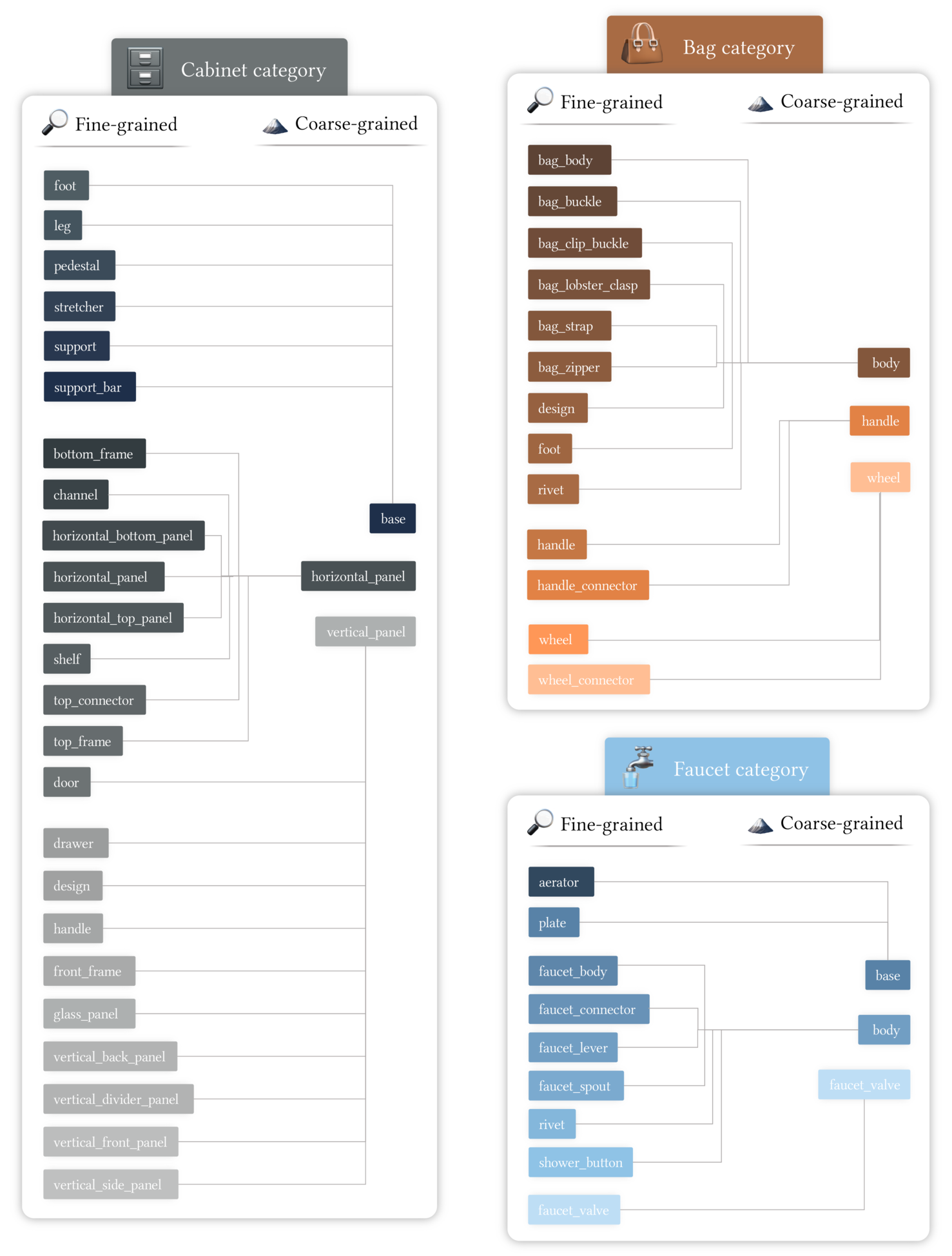}
  \vspace{1em}
  \caption{\textbf{From fine-grained to coarse-grained segmentation.}
  We consider shapes from the "cabinet", "bag" and "faucet" categories.
  Fine-grained classes are merged following a shape category-specific nomenclature to create coarse-grained classes.
  Resulting shapes are simplified and contain fewer parts.}
  \label{fig:multi_levels__page}
\end{figure*}

Additional examples for fine-grained/coarse-grained part and material part categories groupings are provided in Figure \ref{fig:multi_levels__page}.

\section{}
\begin{figure*}
  \centering
  \includegraphics[width=\linewidth]{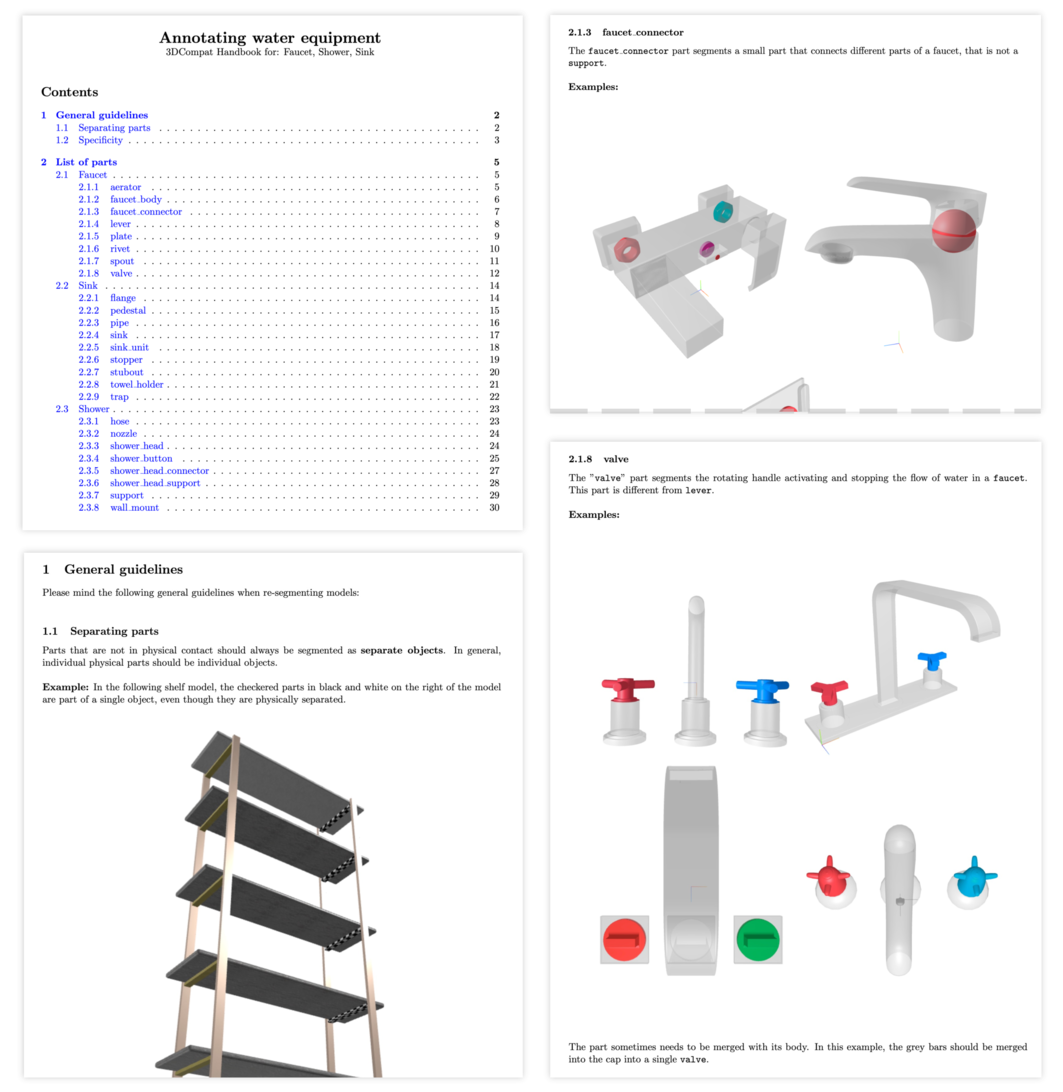}

  \vspace{1em}
  \caption{
    \textbf{Excerpts from the \compat guidelines.}
    We show excerpts from the \compat guidelines for the \texttt{faucet, shower, sink} shape categories.
    Related shape categories with a large number of shared parts are defined jointly in the guidelines.
    In total, 33 annotation guidelines were created and refined for the \nshapeclasses shape categories of \compat,
    for a total of 606 pages of documentation.
  }
  \label{fig:guideline_short}
\end{figure*}

Examples for guidelines provided to annotators for the part segmentation task are provided in Figure \ref{fig:guideline_short}.

\noindent We provide all guidelines PDFs for reference in the following link: \url{https://3dcompat-dataset.org/v2/guidelines}.

\begin{figure*}[hbtp!]
  \vspace{1em}
  \centering
  {
  \includegraphics[width=\linewidth]{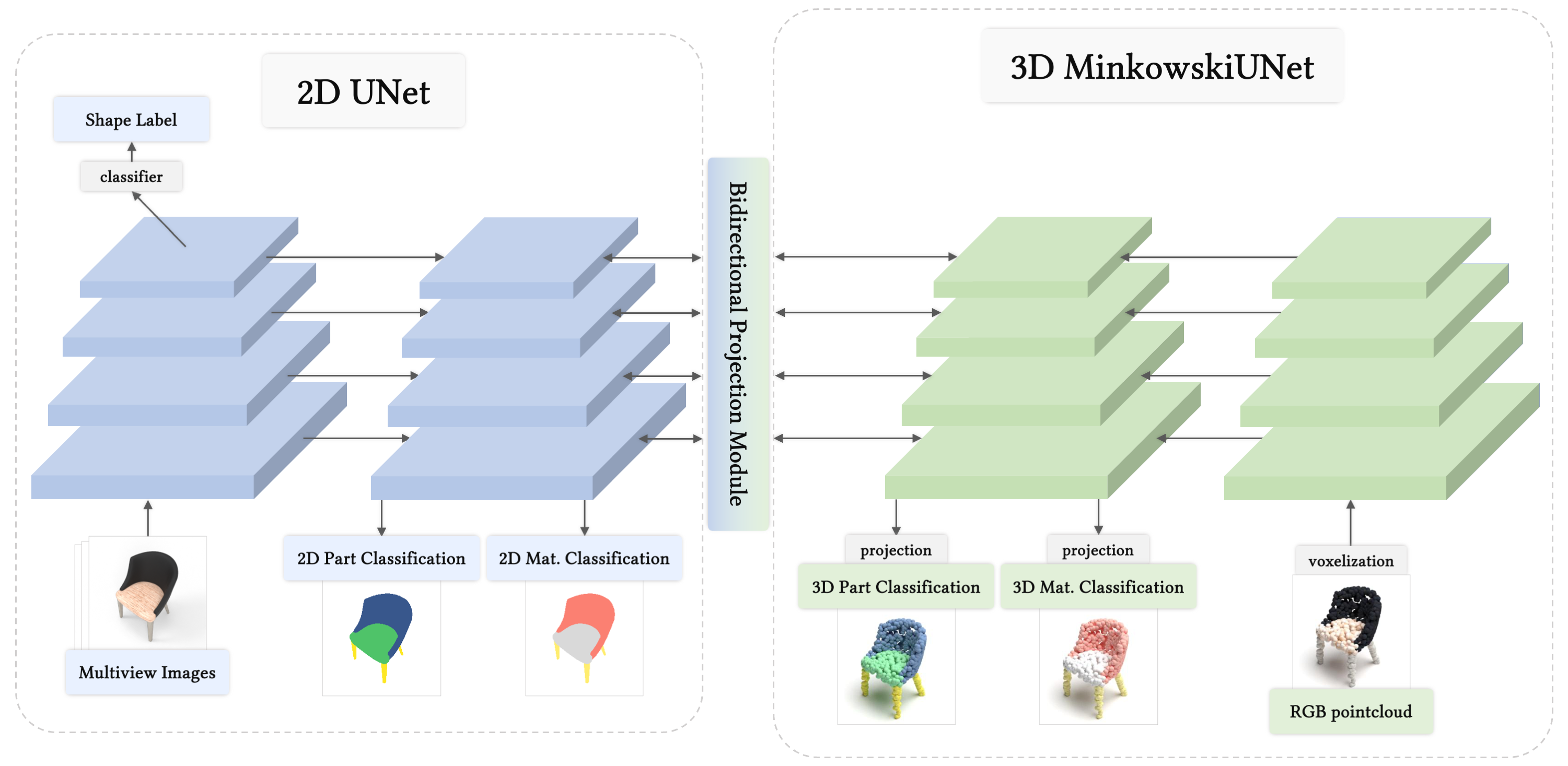}
  }
  \vspace{-1em}
  \caption{
    \textbf{Modified BPNet architecture details.}
    We provide here additional details about the modified BPNet~\cite{hu_bpnet_2021} architecture we employ for the GCR task.
    We extract shape classification logits from the 2D UNet branch, and part-material logits from both the 2D UNet and the 3D MinkowskiNet branches. We use voxelized pointclouds as input to the 3D branch and multi-view renderings as inputs to the 2D branch.
    3D logits are back-projected to the 3D pointclouds to produce part-material predictions in the 3D space.
  }
  \label{fig:bpnet_architecture}
\end{figure*}

\begin{figure*}[hbtp!]
  \vspace{1em}
  \centering
  {
  \includegraphics[width=\linewidth]{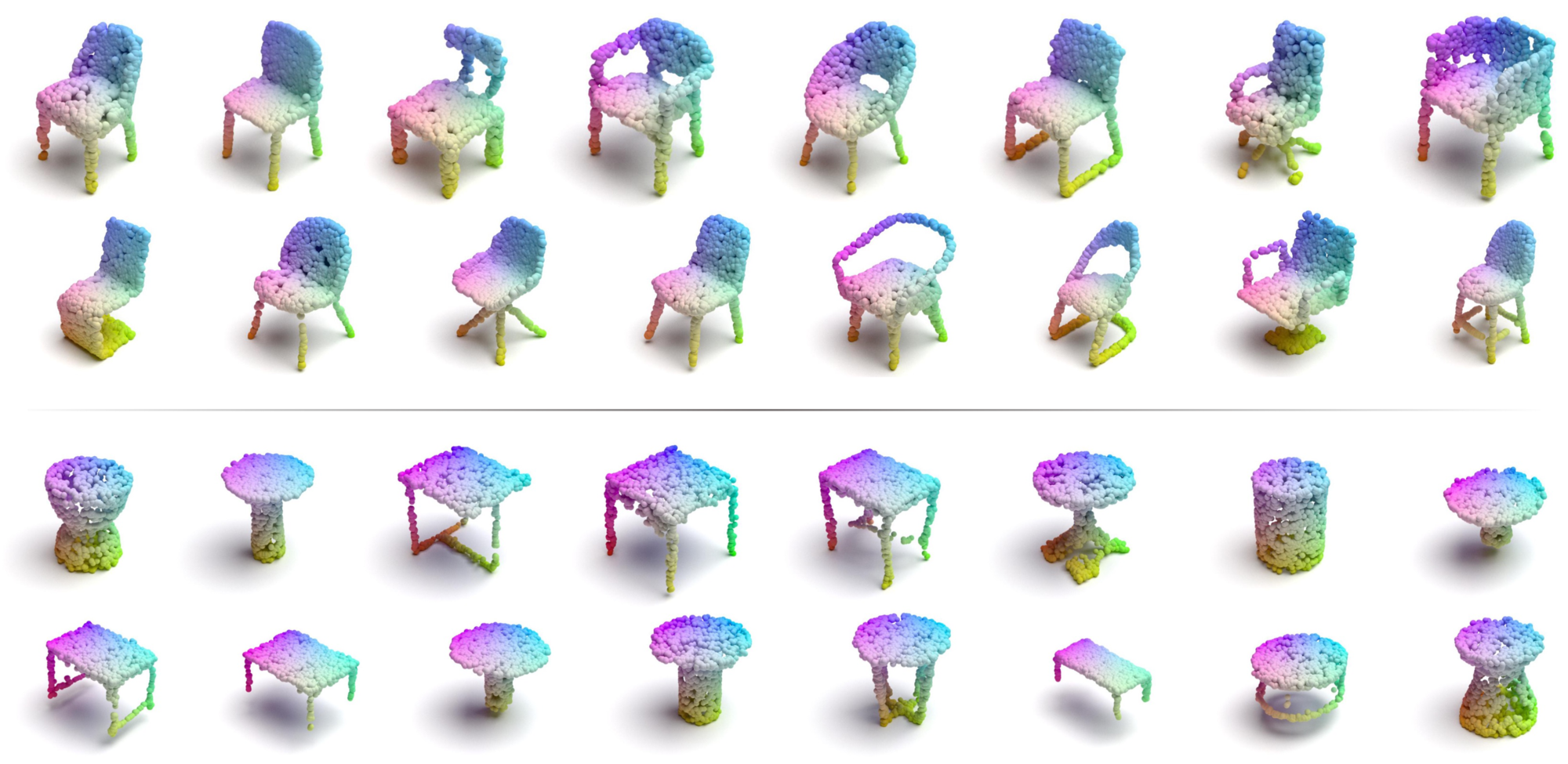}
  }
  \caption{
    \textbf{Generation examples with \compat shapes.}
    Examples of LION~\cite{zeng_lion_2022} generated shapes for chairs (\textbf{top}) and tables (\textbf{bottom}).
    We train LION from scratch using sampled pointclouds from \compat.
    Generated shapes are diverse and visually plausible.
  }
  \label{fig:lion_gen}
\end{figure*}

\section{}
We conduct shape generation experiments on \compat on the \texttt{table} and \texttt{chair} categories.
We use LION~\cite{zeng_lion_2022} to generate shapes as pointclouds, with models trained separately for each category.
LION is a latent diffusion denoising diffusion model (DDM~\cite{ho_denoising_2020}) that learns a hierarchical latent space of point clouds.
In Figure \ref{fig:lion_gen}, we show examples of generated shapes for both categories.
Overall, generated shapes are diverse and realistic, highlighting the potential of \compat for more general shape comprehension and generation tasks.

\section{}
Detailed architecture of the modified BPNet~\cite{hu_bpnet_2021} model we employ for the GCR task is provided in Figure \ref{fig:bpnet_architecture}.

\clearpage

\bibliographystyle{ieeetr}
\bibliography{citations}

\newpage
\section*{Biography Section}

\vspace{-2em}

\begin{IEEEbiography}[{
  \includegraphics[width=1in,height=1.15in,clip]{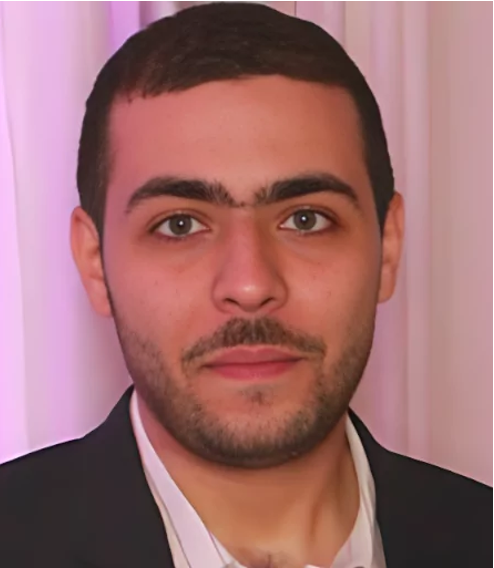}}
  \vspace{0em}]{Habib Slim}
  is a Ph.D. student at KAUST, Saudi Arabia.
  He earned a M.Res. in Data Science from Université Grenoble Alpes (UGA), France, during which he worked on class-incremental learning for image classification at Université Paris-Saclay.
  He received a M.Eng. in Computer Science from École Nationale Supérieure d'Informatique et de Mathématiques Appliquées de Grenoble (ENSIMAG) in 2020.
  He is interested in continual/compositional 2D/3D vision.
\end{IEEEbiography}

\vspace{-3em}

\begin{IEEEbiography}[{
  \includegraphics[width=1in,height=1.15in,clip]{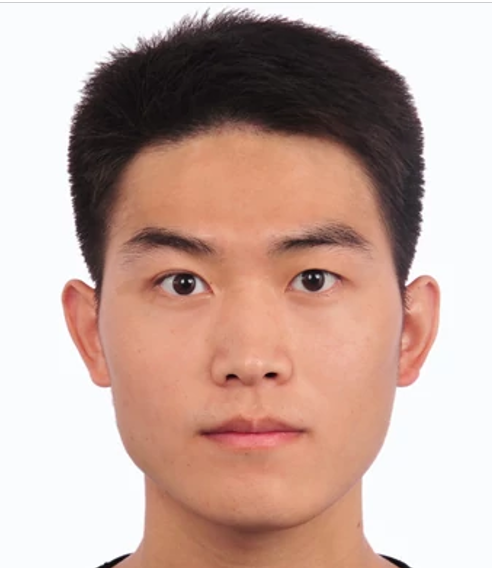}}
  \vspace{0em}]{Xiang Li} is a postdoctoral researcher in computer vision at KAUST, Saudi Arabia.
  He received a B.S. degree in Remote Sensing Science and Technology from Wuhan University, Wuhan, China, in 2014.
  He received a Ph.D. in Cartography and GIS from the Institute of Remote Sensing and Digital Earth, Chinese Academy of Sciences, China, in 2019.
  His research interests include computer vision, deep learning, and remote sensing.
\end{IEEEbiography}

\vspace{-3em}

\begin{IEEEbiography}[{
  \includegraphics[width=1in,height=1.15in,clip]{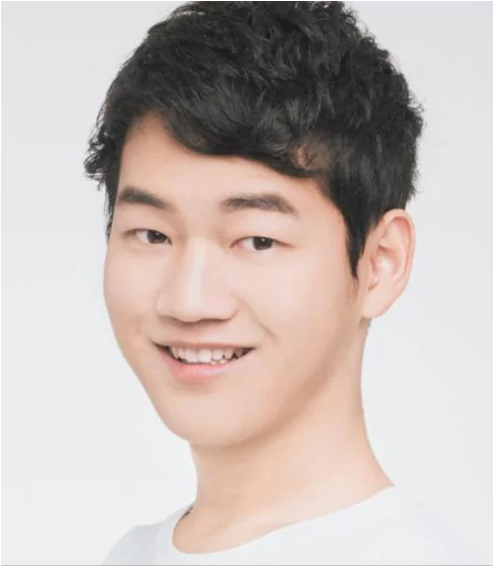}}
  \vspace{0em}]{Yuchen Li}
  is a PhD Student at KAUST, Saudi Arabia. 
  Before joining KAUST, Yuchen was a research intern at iFLYTEK, an intelligent speech and artificial intelligence company in Hefei, China, for three months.
  He was a Rocket MQ open source contributor and Alibaba summer of code student developer with rich research and engineering experience. 
  He is interested in meta-learning, few-shot learning and 3D object recognition.
\end{IEEEbiography}

\vspace{-3em}

\begin{IEEEbiography}[{
  \includegraphics[width=1in,height=1.15in,clip]{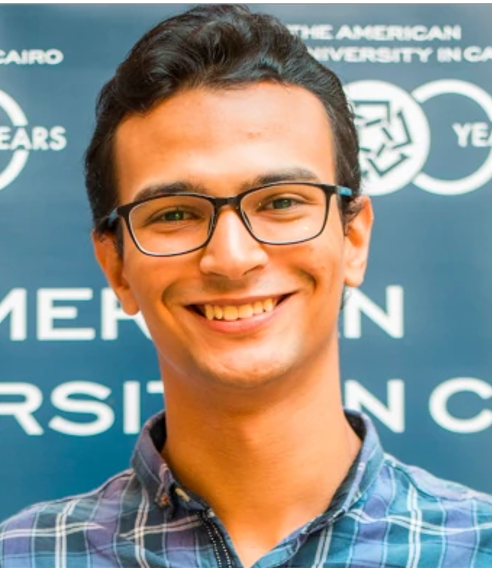}}
  \vspace{0em}]{Mahmoud Ahmed}
  is an M.S student at KAUST, Saudi Arabia, and received his B.S degree from the American University in Cairo (AUC), Egypt, in 2022.
  Prior to that, he worked as a Data Science intern at Dell Technologies, then as a 5G Software Engineer.
  His research interests include computer vision, graphics, and deep learning.
\end{IEEEbiography}

\vspace{-3em}

\begin{IEEEbiography}[{
  \includegraphics[width=1in,height=1.15in,clip]{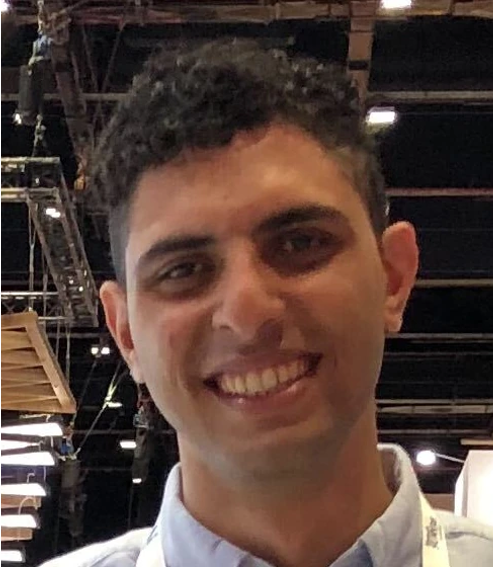}}
  \vspace{0em}]{Mohamed Ayman}
  is a M.S. student at University of Alberta and received his B.S. degree from the American University in Cairo  (AUC), Egypt, in 2023.
  He worked as an Applied Science intern at Microsoft.
  Currently, he is a research intern at KAUST, Saudi Arabia.
  His research interests are focused on computer vision, NLP, and software optimization.
\end{IEEEbiography}

\vspace{-3em}

\begin{IEEEbiography}[{
  \includegraphics[width=1in,height=1.15in,clip]{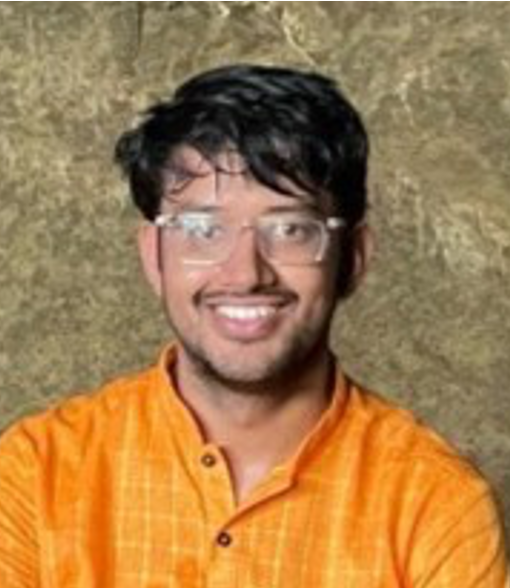}}
  \vspace{0em}]{Ujjwal Upadhyay} is an AI Scientist at qure.ai where he works on applying novel deep learning methods to medical data.
  His research interests include computer vision, adversarial machine learning, and representation learning.
  He has been involved in cutting-edge research in 3D vision, scene understanding, and neuroscience.
\end{IEEEbiography}

\newpage
{\color{white}
-
}
\vspace{-0.55em}
\begin{IEEEbiography}[{
  \includegraphics[width=1in,height=1.15in,clip]{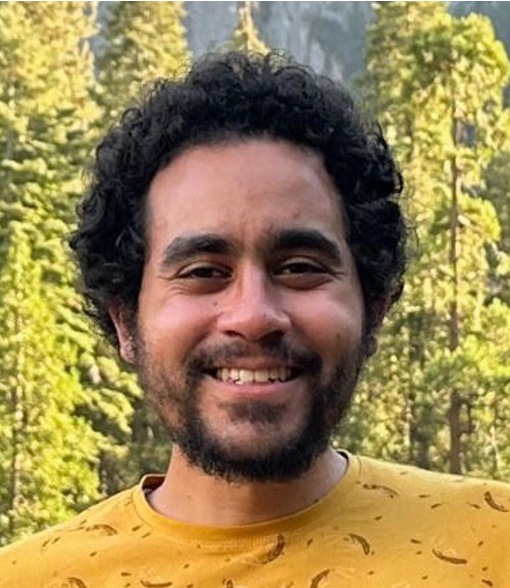}}
  \vspace{0em}]{Ahmed Abdelreheem}
  is a Ph.D. student at KAUST, Saudi Arabia.
  He received a BSc in Computer Engineering from Cairo University, Egypt, in 2019.
  He attained his MSc degree in Computer Science from KAUST, Saudi Arabia, in 2022.
  His research interests lie in the intersection of 3D vision, computer graphics, and natural language.
  More specifically, he is interested in linking 3D object-centric representations to natural language.
\end{IEEEbiography}

\vspace{-1.15em}

\begin{IEEEbiography}[{
  \includegraphics[width=1in,height=1.15in,clip]{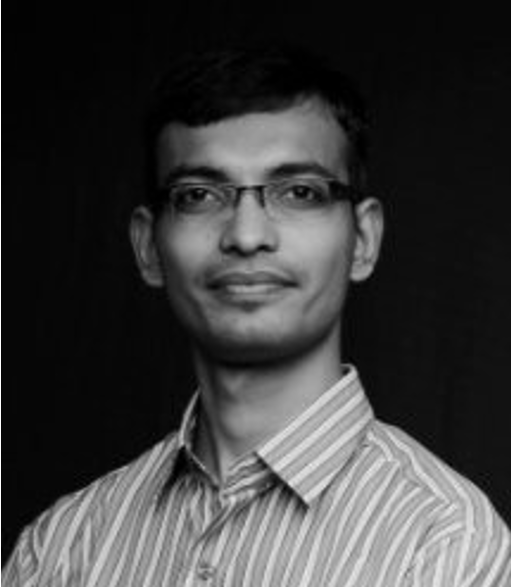}}
  \vspace{0em}]{Arpit Prajapati}
  is the Director of Technology at Poly9, where his responsibilities encompass developing solutions for product sampling in the Home and Lifestyle industry.
  Prior to this role, he was the owner of Lanover Solutions for nearly 8 years, managing company operations.
  He holds a Bachelor of Engineering (BE) degree in Computer from Gujarat University, completed between 2005 and 2009.
\end{IEEEbiography}

\vspace{-1.95em}

\begin{IEEEbiography}[{
  \includegraphics[width=1in,height=1.15in,clip]{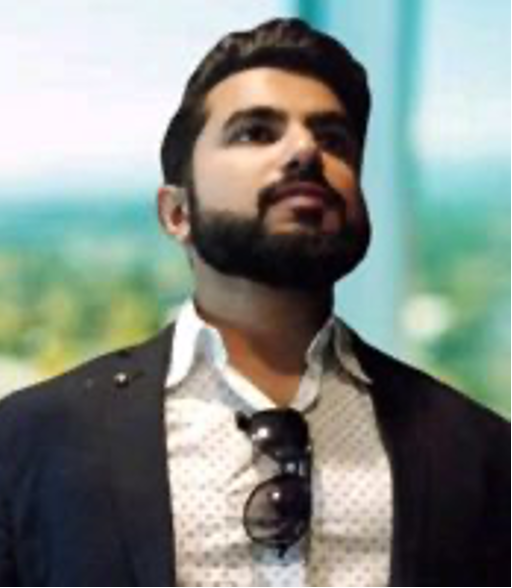}}
  \vspace{0em}]{Suhail Pothigara} is the CEO of Poly9, a recognized Product Management leader in ecommerce and digital transformation.
  With 15 years of experience at luxury brands and retailers in fashion, home, and consumer electronics, his accomplishments include driving over \$3 billion in revenue cumulatively through his roles at e-commerce and cloud businesses at Macy's, LVMH, and HP.
\end{IEEEbiography}

\vspace{-2em}

\begin{IEEEbiography}[{
  {
    \includegraphics[width=1in,height=1.15in,clip]{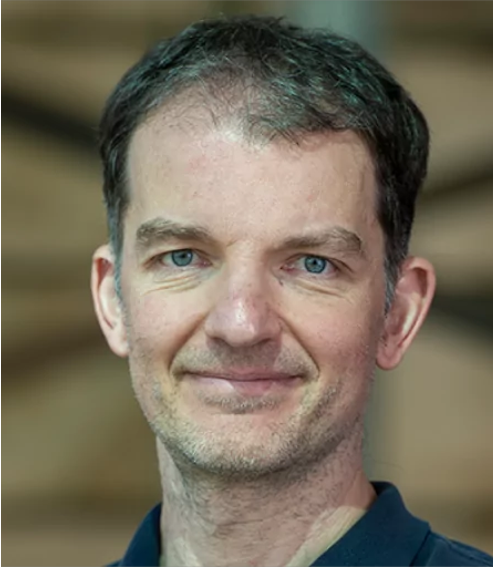}}
    \vspace{-0.5em}
  }
  \vspace{0em}]{Peter Wonka}

  is Full Professor in Computer Science at KAUST, Saudi Arabia, and Interim Director of the Visual Computing Center (VCC). Peter Wonka received his Ph.D. from the Technical University of Vienna in computer science.
  Additionally, he received a M.Sc. in Urban Planning from the same institution.
  After his Ph.D., he worked as a postdoctoral researcher at the Georgia Institute of Technology and as faculty at Arizona State University.
  His research publications tackle various topics in computer vision, computer graphics, remote sensing, image processing, visualization, and machine learning.
  The current research focus is on deep learning, generative models, and 3D shape analysis and reconstruction.
\end{IEEEbiography}

\vspace{-2em}

\begin{IEEEbiography}[{
  {
    \includegraphics[width=1in,height=1.15in,clip]{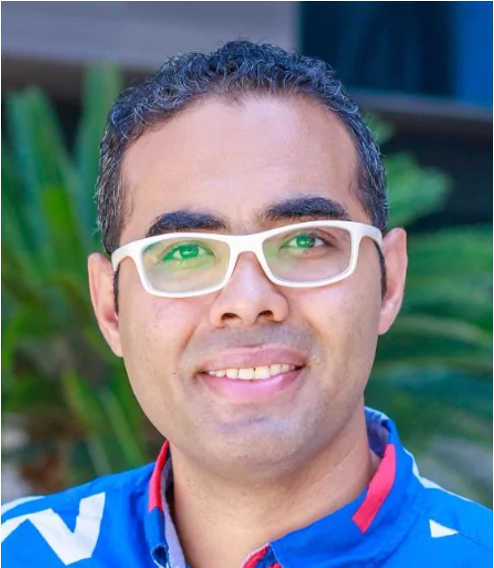}
    \vspace{-1em}
  }
  }
  \vspace{0em}]{Mohamed Elhoseiny}
is an Assistant Professor of Computer Science at KAUST, Saudi Arabia, and a Senior Member of IEEE and AAAI.
He was a Visiting Faculty at Stanford Computer Science Department (2019-2020), a Visiting Faculty at Baidu Research Silicon Valley Lab (2019), and a Postdoc Researcher at Facebook AI Research (2016-2019).
Dr. Elhoseiny completed his Ph.D. in 2016 at Rutgers University, during which he spent time at Adobe Research (2015-2016) for more than a year and at SRI International in 2014.
He received an NSF Fellowship in 2014 and the Doctoral Consortium Award at CVPR 2016.
His primary research interest is in computer vision, especially in efficient multimodal learning with limited data in areas like zero/few-shot learning, vision and language, and language-guided visual perception.
He is also interested in affective AI and particularly in producing novel art and fashion with AI. His creative AI work was featured in MIT Tech Review, New Scientist Magazine, and the HBO show Silicon Valley.
\end{IEEEbiography}
\vfill

\end{document}